\documentclass[lettersize,journal]{IEEEtran}
\usepackage{amsmath,amsfonts}
\usepackage{algorithmic}
\usepackage{algorithm}
\usepackage{array}
\usepackage[caption=false,font=normalsize,labelfont=sf,textfont=sf]{subfig}
\usepackage{textcomp}
\usepackage{stfloats}
\usepackage{url}
\usepackage{verbatim}
\usepackage{graphicx}
\usepackage{cite}
\usepackage[table,xcdraw]{xcolor}
\usepackage[hidelinks]{hyperref}       % hyperlinks
\usepackage{booktabs}       % professional-quality tables
\usepackage{nicefrac}       % compact symbols for 1/2, etc.
\usepackage{microtype}      % microtypography
\usepackage[T1]{fontenc}
\usepackage{amssymb}
\usepackage{makecell}
\usepackage{multirow}
\usepackage{pifont}

\newcommand{\ie}{\textit{i}.\textit{e}.}
\newcommand{\eg}{\textit{e}.\textit{g}.}
\usepackage{arydshln}
\DeclareMathOperator*{\argmax}{argmax}
\usepackage{lscape}

\begin{document}

\title{Textless Unit-to-Unit training for Many-to-Many Multilingual Speech-to-Speech Translation}

\author{Minsu Kim$^{\dagger}$, Jeongsoo Choi$^{\dagger}$, Dahun Kim, Yong Man Ro,~\IEEEmembership{Senior Member,~IEEE}% <-this % stops a space
\thanks{This work is partially supported by Institute of Information \& communications Technology Planning \& Evaluation (IITP) grant funded by the Korea government (MSIT) (No.2022-0-00124, Development of Artificial Intelligence Technology for Self-Improving Competency-Aware Learning Capabilities) and the National Research Foundation of Korea (NRF) grant funded by the Korea government (MSIT) (No.~NRF-2022R1A2C2005529)}
\thanks{M. Kim, J. Choi, and Y. M. Ro are with the Integrated Vision and Language Lab., School of Electrical Engineering, Korea Advanced Institute of Science and Technology (KAIST), 291 Daehak-ro, Yuseong-gu, Daejeon, 34141, Republic of Korea (e-mail: ms.k@kaist.ac.kr; jeongsoo.choi@kaist.ac.kr; ymro@kaist.ac.kr). D. Kim is with Google DeepMind, Mountain View, CA 94043, USA (e-mail: mcahny01@gmail.com). Corresponding author: Y. M. Ro (fax: 82-42-350-5494). $^{\dagger}$Both authors contributed equally to this work.}% <-this % stops a space
}


% % column for its text to clear the IEEEpubid mark.

\maketitle

\begin{abstract}
This paper proposes a textless training method for many-to-many multilingual speech-to-speech translation that can also benefit the transfer of pre-trained knowledge to text-based systems, text-to-speech synthesis and text-to-speech translation. To this end, we represent multilingual speech with speech units that are the discretized representations of speech features derived from a self-supervised speech model. By treating the speech units as pseudo-text, we can focus on the linguistic content of the speech, which can be easily associated with both speech and text modalities at the phonetic level information. By setting both the inputs and outputs of our learning problem as speech units, we propose to train an encoder-decoder model in a many-to-many spoken language translation setting, namely Unit-to-Unit Translation (UTUT). Specifically, the encoder is conditioned on the source language token to correctly understand the input spoken language, while the decoder is conditioned on the target language token to generate the translated speech in the target language. Therefore, during the training, the model can build the knowledge of how languages are comprehended and how to relate them to different languages. Since speech units can be easily associated from both audio and text by quantization and phonemization respectively, the trained model can easily transferred to text-related tasks, even if it is trained in a textless manner. We demonstrate that the proposed UTUT model can be effectively utilized not only for Speech-to-Speech Translation (S2ST) but also for multilingual Text-to-Speech Synthesis (T2S) and Text-to-Speech Translation (T2ST), requiring only minimal fine-tuning steps on text inputs. By conducting comprehensive experiments encompassing various languages, we validate the efficacy of the proposed method across diverse multilingual tasks. Moreover, thanks to the many-to-many language training, we show that the UTUT can also perform language translations for novel language pairs that are not present during training as pairs, which has not well been explored in the previous literature. Samples can be found on \url{https://choijeongsoo.github.io/utut}.
\end{abstract}


\section{Introduction}

\IEEEPARstart{E}{ven} though we grew up in different countries, we can easily communicate with people having different nationalities with recent advanced technologies, Automated Speech Recognition (ASR) \cite{hannun2014deepspeech,amodei2016deep,kim2017joint,watanabe2017hybrid,watanabe2018espnet,kim2021cromm,shi2022avhubert,kim2022distinguishing,hong2023watch}, Neural Machine Translation (NMT) \cite{bahdanau2014neural,wu2016google,chen2018best,brown2020gpt-3,liu2020mbart}, and Text-to-Speech Synthesis (T2S) \cite{wang2017tacotron,jia2018transfer,valle2020flowtron,chen2021adaspeech,casanova2022yourtts,wang2023neural}. As the demand for communication tools with foreigners continues to grow, technologies are being developed to support multilingualism, reflecting the trend of globalization. One of the most popular large language models \cite{thoppilan2022lamda,touvron2023llama}, Chat-GPT \cite{chatgpt} has been trained in over 60 languages and is successfully providing services, including NMT.

However, the technologies handling speech, especially speech synthesis technologies, are encountering challenges in being extended to multilingual \cite{zhang2019learning,nekvinda2020one,zhang2023valle-x,saeki2023learning}. They are typically developed to handle only one language (\ie, monolingual), so different models should be prepared to apply them to different languages. The main impediment in building multilingual models is the inherent complexity of speech data. While the text is naturally discrete and only covers linguistic content, the speech is continuous and conveys various speaker characteristics such as voice, accent, and timbre. This inherent characteristic of speech puts a challenge in focusing only on the linguistic content of multilingual speech, while the text from different languages can be easily handled with a fixed vocabulary size by using subword tokenizers \cite{sennrich2015bytepair,kudo2018subword,kudo2018sentencepiece} and phonemizers \cite{zhang2019learning,Bernard2021}.

In recent days, speech unit \cite{lakhotia2021generative}, a discretized speech feature, has shown remarkable potential in speech synthesis \cite{lakhotia2021generative,polyak2021speech,choi2023intelligible} and speech recognition \cite{shi2022avhubert,hsu2021hubert,chang2023exploration}. These speech units, closely related to acoustic units such as phonemes \cite{hsu2021hubert}, are obtained by quantizing (\ie, clustering) extracted speech representations derived from a self-supervised speech model \cite{hsu2021hubert,baevski2020wav2vec,babu2021xls}. By processing speech using speech units, a fascinating possibility arises: treating speech units as pseudo text, thus enabling textless Natural Language Processing (NLP) \cite{lakhotia2021generative,borsos2023audiolm}.
As different languages share some common phonemes \cite{schultz2001multilingualASR,vu2014multilingual,luo2020multilingualVSR}, it is available to tokenize multilingual speech with a fixed vocabulary size by using speech units (\ie, the acoustic units), similar to phonemizers in text modality. Moreover, the speech units enable the establishment of a straightforward connection between speech and text at the phonetic level, thereby facilitating the seamless transfer of knowledge between these two modalities.

In this paper, by employing the characteristics of speech units, we propose a textless many-to-many multilingual speech-to-speech translation method, namely Unit-to-Unit Translation (UTUT). Since the proposed method is textless, the developed machine translation system can be applied even for languages that have no writing systems \cite{lee2021textless}. To this end, UTUT training proceeds by setting the inputs and outputs of a model with speech units derived from multilingual speech. Its learning problem is composed of a many-to-many spoken language translation problem by conditioning the encoder with the source language token and the decoder with the target language token. Through the many-to-many language UTUT training, the model can effectively build knowledge on how languages are comprehended and how to convert them into various other languages, even with much smaller amounts of data compared to the unidirectional model. By associating the speech units with speech or text, the trained model can be easily transferred to diverse speech- and text-related downstream tasks, Speech-to-Speech Translation (S2ST), multilingual Text-to-Speech Synthesis (T2S), and Text-to-Speech Translation (T2ST), even though the model is originally trained without text.

The major contributions of this paper are as follows,
\begin{itemize}
    \item We propose a textless Unit-to-Unit training method for many-to-many multilingual speech-to-speech translation. As the speech units mainly contain the linguistic information ($\eg$, phoneme), the trained model can be easily transferred to text-based systems of Text-to-Speech synthesis (T2S) and Text-to-Speech Translation (T2ST), despite being trained in a text-free manner.
    \item To the best of our knowledge, this is the first work exploring textless and many-to-many language translation in S2ST. By training the model with the proposed Unit-to-Unit Translation (UTUT), we can effectively build S2ST systems even with much smaller training data compared to unidirectional S2ST models.
    \item We show that through the proposed training with many-to-many languages, we can perform language translations even for novel language pairs that are never given as pairs during training.
\end{itemize}

\section{Related Work}
\textbf{Self-Supervised Speech Model.}
Pre-trained speech processing models using self-supervised learning have achieved significant performances in ASR \cite{baevski2020wav2vec,hsu2021hubert}, speaker verification \cite{chen2022wavlm}, and speech synthesis \cite{lakhotia2021generative,ao2021speecht5}. HuBERT \cite{hsu2021hubert} proposed a self-supervised learning method that predicts hidden units obtained by clustering the acoustic features (\eg, MFCC) from masked audio. By refining the target hidden units with clusters of the trained features, they achieved state-of-the-art speech recognition performances even in the low-resource setups. AV-HuBERT \cite{shi2022avhubert} successfully extended the training method of HuBERT to audio-visual speech representation learning, and even showed powerful lip reading \cite{chung2017lip,hong2021speech,kim2023prompt,yeo2023multi} performances by finetuning the trained model on the labeled dataset.

Textless Natural Language Processing (NLP) has emerged with high-quality speech representations extracted by self-supervised speech models \cite{baevski2020wav2vec,hsu2021hubert}. By quantizing the speech features, they obtained speech units that can be utilized as pseudo texts, where non-linguistic features are suppressed and linguistic contents are kept \cite{chang2023exploring}. By treating the speech units as pseudo texts, spoken language modeling \cite{lakhotia2021generative,nguyen2023generative}, speech emotion conversion \cite{kreuk2021textlessemotion}, and speech resynthesis \cite{polyak2021speech} are proposed.

Based on the speech units obtained by quantizing multilingual speech, in this paper, we try to build textless many-to-many multilingual S2ST model. Since speech units are highly correlated with acoustic units, we can easily transfer the trained model to diverse multilingual speech- and text-related downstream tasks.

\textbf{Speech-to-Speech Translation.}
Speech-to-Speech Translation (S2ST) \cite{do2016preserving} aims to directly synthesize the translated speech in the target language from the input speech of the source language. Recently, the technology has drawn big attention, since it can reduce error propagation and time consumption of cascaded systems of ASR \cite{li2014overview,abdel2014convolutional,zhang2020improving}, MT \cite{su2018hierarchy,yang2022gtrans}, and T2S \cite{huang2022meta,du2023speaker}. Early methods tried to synthesize spectrograms from input speech in an end-to-end manner. Translatotron \cite{jia2019translatotron} proposed a spectrogram-to-spectrogram model trained with auxiliary ASR and Speech-to-Text Translation \cite{kano2020end} losses by using text supervision. Translatotron2 \cite{jia2021translatotron2}, the enhanced model of the previous version, proposed to jointly train the model with a speech-to-speech translation objective and a speech-to-phoneme translation objective. \cite{kano2021transformer} proposed to use Transformer sequence-to-sequence model \cite{vaswani2017attention} in S2ST. 

As the speech unit can be treated as pseudo text, speech-to-unit translation models are proposed instead of predicting spectrograms \cite{huang2022transpeech,inaguma2022unity}. \cite{lee2021directs2s} used speech units obtained from HuBERT \cite{hsu2021hubert} as targets and trained the model with multitask learning of text and speech unit prediction. \cite{lee2021textless} proposed a textless S2ST model which does not utilize text supervision. \cite{jia2022leveraging} showed using a pre-trained speech encoder, w2v-BERT\cite{chung2021w2v}, can boost the S2ST performance. \cite{popuri2022enhanced} proposed to use a pre-trained speech encoder and a pre-trained unit decoder, and proposed finetuning strategies. \cite{diwan2024textless} extended the previous work to operate with low-resource languages by learning with a round-trip translation consistency loss, which uses the back translation \cite{sennrich2016improving}. Recently, \cite{barrault2023seamlessm4t} extended \cite{inaguma2022unity} with multimodal and large-scale datasets and proposed a powerful multimodal and multilingual machine translation system.

In this paper, we train the model with a Unit-to-Unit Translation (UTUT) task constructed as a many-to-many spoken language translation, without using text supervision. Through many-to-many language translation learning, the proposed method can effectively learn with significantly fewer parallel data compared to language-specific translation models. Moreover, the proposed UTUT can perform language translation between novel language pairs for which parallel data is absent in the training set. The proposed method can be easily extended to text-related tasks, Text-to-Speech Translation (T2ST) and multilingual Text-to-Speech Synthesis (T2S). Please note that different from the recently proposed method \cite{barrault2023seamlessm4t}, the proposed method is a textless training method and is trained in a many-to-many spoken language translation setting, while their model is trained using Many-to-En and En-to-Many translations.

\textbf{Multilingual Text-to-Speech Synthesis.}
Text-to-Speech Synthesis (T2S) is one of the well-studied research subjects \cite{wang2017tacotron,jia2018transfer,ren2019fastspeech,arik2017deep,cooper2020zero}. It enables computers to speak like humans from the written text. However, it is not easy to develop the technology for languages with low-resource data and for handling multilingual data. \cite{zhang2019learning} proposed to use shared phonemic input across different languages and to disentangle the speaker characteristics and content. \cite{azizah2020hierarchical,xu2020lrspeech} proposed a method that works in low-resource scenarios by using transfer learning from multilingual data. \cite{he2021multilingual} trained a T2S model on multilingual data and showed that the multilingual trained model can learn a new language with a few training samples. \cite{lux2022low} proposed to use language-agnostic meta-learning with modified encoder architecture for enabling few-shot adaptation to a new language. YourTTS \cite{casanova2022yourtts} firstly explored zero-shot speaker transfer on multilingual T2S. By injecting a language token and speaker embeddings, they appropriately synthesized multilingual speech with the desired target speaker's voice. VALL-E X \cite{zhang2023valle-x} proposed to train a neural codec language model by using EnCodec \cite{defossez2022high} for speech representation. By learning a neural codec language model, they can synthesize cross-lingual speech reducing the foreign accent problem.

We train the model with multilingual speech-only data without text through a UTUT objective. Since the speech unit is strongly tied to the phonemes, the trained model can be easily transferred for multilingual T2S. Moreover, by setting the target language differently from the source language, we can also perform multilingual Text-to-Speech Translation (T2ST) with a single model. Please note that this paper is a pioneering work exploring many-to-many multilingual T2ST.

\section{Proposed Method}
Let $x\in \mathbb{R}^{T \times C}$ be a speech with length $T$ and channel size of $C$. Then, we quantize the speech into speech units $u\in \mathbb{R}^{S}$ by clustering the extracted features from a self-supervised speech model, where $S$ is the length of the speech units. The length of the speech units depends on how much the used self-supervised speech model subsamples the input audio $x$. Our objective is to learn multilingual speech representations without using text. Specifically, we design our learning problem as a Unit-to-Unit Translation (UTUT) constructed as a many-to-many spoken language translation so that the speech units in the source language are translated into those of the desired target language.

%------------------------------------ Figure 1
%#############################################
\begin{figure*}[t]
	\begin{minipage}[b]{1.0\linewidth}
		\centering
		\centerline{\includegraphics[width=13.5cm]{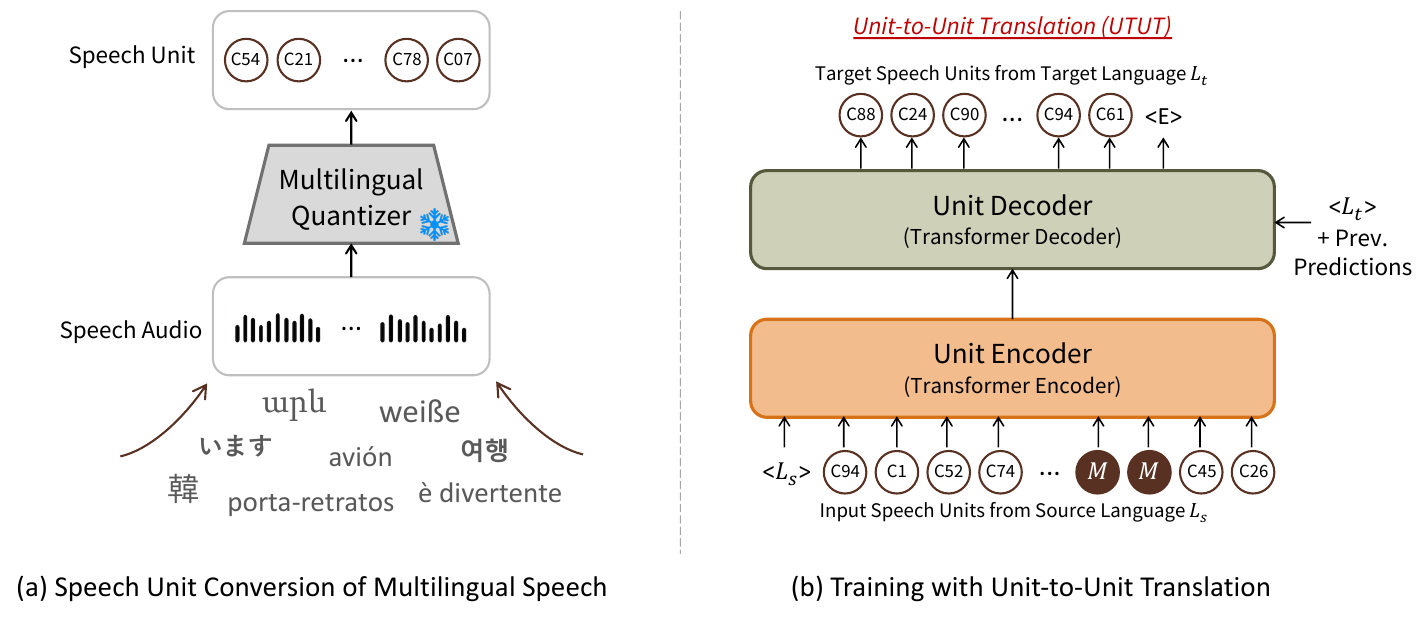}}
	\end{minipage}
	\caption{Overview of the proposed UTUT framework. (a) Since different languages can be represented with a common phoneme set, it is possible to quantize multilingual speech into speech units with a fixed vocabulary size. (b) The model is trained with a unit-to-unit translation objective by setting the model's inputs and outputs as speech units from a source and a target language.}
	\label{fig:1}
\end{figure*}
%#############################################

\subsection{Speech Unit Extraction}
Recently, textless NLP technologies \cite{lakhotia2021generative,nguyen2023generative} successfully learned a spoken language model by discretizing the input speech into speech units. They showed that by quantizing audio features extracted by a pre-trained self-supervised speech model such as wav2vec2.0 \cite{baevski2020wav2vec} and HuBERT \cite{hsu2021hubert}, we can focus on the linguistic content of speech while suppressing the other factors. Motivated by the recent success of textless NLP, we embed multilingual speech into speech units that are closely correlated with acoustic units \cite{hsu2021hubert,sicherman2023analysing}. Since different languages can be expressed by a common phoneme vocabulary set \cite{schultz2001multilingualASR,vu2014multilingual,luo2020multilingualVSR}, we can tokenize the multilingual speech into speech units with a fixed dictionary size. In addition, by processing with the speech units, we can learn the joint representations of speech and text which can be easily transferred to multimodalities \cite{zhang2022speechut}. The process for converting speech into speech units is illustrated in Fig.\ref{fig:1}a.

For the pre-trained self-supervised speech model, we employ multilingual HuBERT (mHuBERT) \cite{hsu2021hubert,li2022textlesss2s} trained on an unlabeled speech dataset from VoxPopuli \cite{wang2021voxpopuli}. The model takes 16kHz and 1 channel audio $x$ as input and embeds it into speech features. The resulting speech features are 50fps. Following \cite{lakhotia2021generative}, we quantize the speech features into units by using K-means clustering and remove sequential repetitions of units, which results in our input and output speech units $u$ for training the model. Following \cite{li2022textlesss2s}, we use the 11-th layer of mHuBERT to extract the speech features and a vocabulary size of 1,000. 

\subsection{Unit-to-Unit Translation (UTUT)}
By quantizing multilingual speech into speech units $u$, we can now process multilingual speech as if it were text defined with a fixed vocabulary size. By treating the speech units as pseudo text, we propose to train our model using a Unit-to-Unit Translation (UTUT) objective. Specifically, we design the learning problem as a many-to-many spoken language translation. We condition the unit encoder with the source language token so that the encoder can know the language of incoming input speech units. Then, from the encoded source language features, the unit decoder generates the target speech units by conditioning on the target language token, in a sequence-to-sequence manner. The training process of UTUT is illustrated in Fig.\ref{fig:1}b.

\subsubsection{Architecture} 
We use standard transformer encoder-decoder architecture \cite{vaswani2017attention} which is composed of 12 transformer encoder layers for the unit encoder and 12 transformer decoder layers for the unit decoder, similar to \cite{lewis2019bart}. The input for the unit encoder is composed of a source language token <$L_s$> and the input speech units. The unit decoder takes a target language token <$L_t$> as a Beginning of Sequence (BOS) token and its previous predictions as inputs, and encoder attention is applied to features encoded by the unit encoder to translate the input speech of the source language into the target language.

\subsubsection{Learning}
\label{sec:3.3.2}
We sample paired S2ST data, $x_s$ and $x_t$, where subscription $s$ represents the speech data from source language $L_s$ and $t$ represents that of the target language $L_t$. Then the speech $x_s$ and $x_t$ are converted into speech units $u_s$ and $u_t$ through the speech unit extraction process, respectively. Then, our learning problem can be represented as follows,
\begin{align}
\label{eq:1}
\argmax_\theta \sum_i \log{p(u_t^i|u_t^{<i},u_s,L_s,L_t;\theta)},
\end{align}
where $\theta$ is the model parameter and $u_t^{<i}$ is the previous predictions before the step $i$. By optimizing Eq. (\ref{eq:1}) for all training samples, the model can comprehend the source spoken language from the input speech units and express it in target speech by producing speech units of the target language. In the context of many-to-many language translation, the training process offers a significant advantage in terms of data efficiency; The model can learn the language from all combinations of language pairs. Hence, much less training data can be employed compared to unidirectional translation training. For instance, the encoder can acquire knowledge of English (En) from all pairs of En-to-X directional translation data, while the decoder can similarly learn En from all pairs of X-to-En data (X indicates multilingual). Therefore, even though the training data does not contain Pt-Es pair, the model still can learn how to translate Pt to Es by using the other pairs of data (\eg, Pt-X and X-Es).

During training, we also utilize the opposite direction of translation data, so that the target to source language data is also used. Moreover, we also examine the effectiveness of input masking strategy \cite{devlin2018bert,lewis2019bart} in UTUT. We randomly mask the input unit sequences. For the masking, $p_m$ percent of the input sequence is masked with sampled spans. The span lengths are drawn from a Poisson distribution ($\lambda$=10) and the frames corresponding to the span are replaced with mask tokens $\textit{M}$.

We train the model for 500k steps with a peak learning rate of 0.003 and warmup steps of 10k. We use Adam optimizer and linear learning rate decay scheduling. Each batch is constructed of up to 1,024 source and target tokens, respectively. All experiments are performed with Fairseq \cite{ott2019fairseq} library.

%------------------------------------ Figure 2
%#############################################
\begin{figure*}[t!]
	\begin{minipage}[b]{1.0\linewidth}
		\centering
		\centerline{\includegraphics[width=14.5cm]{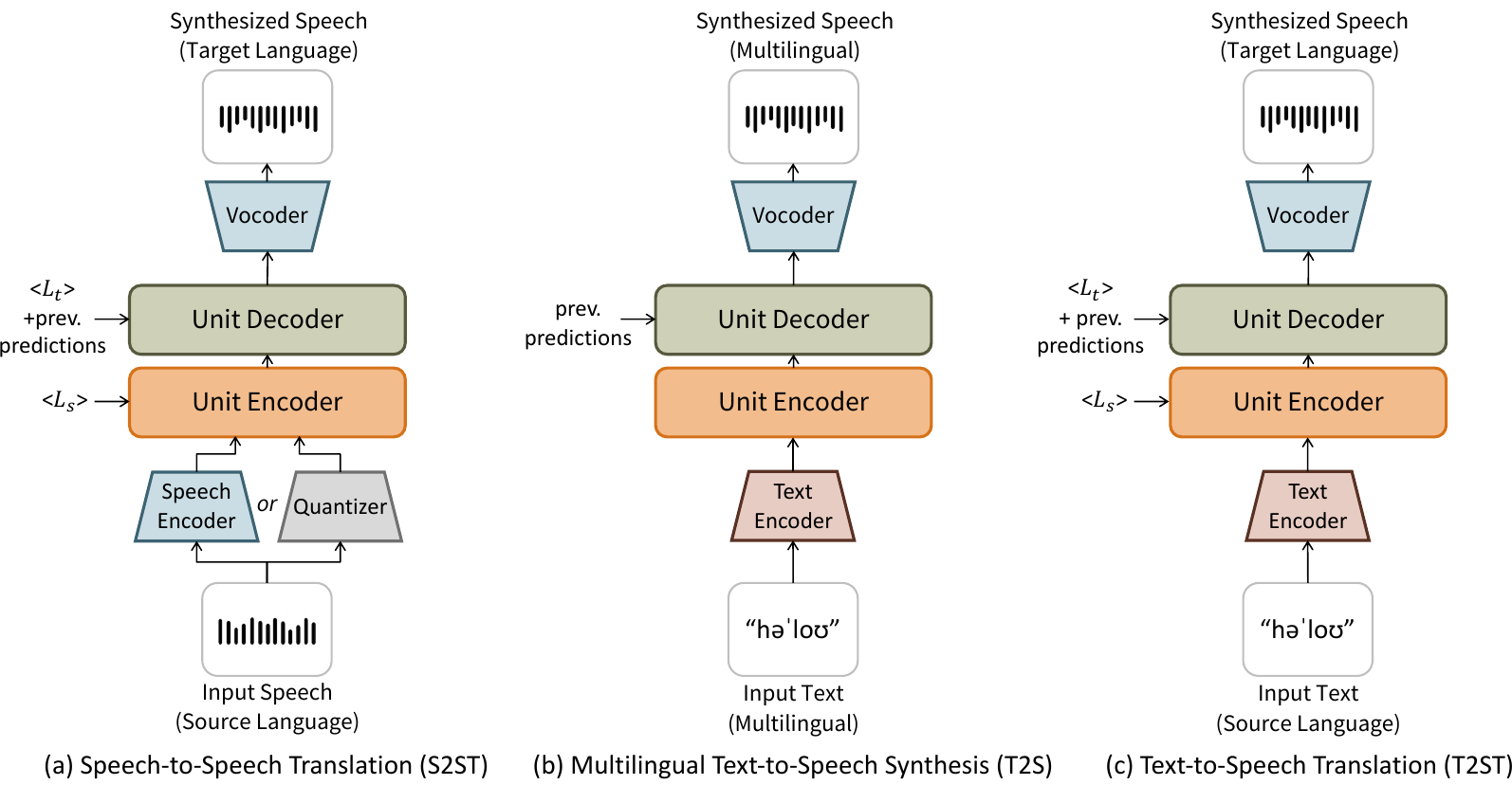}}
	\end{minipage}
	\caption{Example applications of the proposed UTUT framework. (a) Speech-to-Speech Translation (S2ST), (b) Multilingual Text-to-Speech Synthesis (T2S), and (c) Text-to-Speech Translation (T2ST).}
	\label{fig:2}
\end{figure*}
%#############################################

\subsubsection{Downstream Tasks}
The speech units are discrete and mainly contain linguistic information, which means both speech and text can be easily transferred into the learned latent space of speech units \cite{zhang2022speechut}. We show the effectiveness of the proposed UTUT on three different tasks shown in Fig.\ref{fig:2}, Speech-to-Speech Translation (S2ST), multilingual Text-to-Speech Synthesis (T2S), and Text-to-Speech Translation (T2ST). For the task that utilizes speech as input, the speech unit is obtained by using the quantizer that we used or speech is embedded in the trained latent space of the speech unit by using an off-the-shelf speech encoder. For tasks that utilize text as input, we transfer the trained UTUT model by embedding the text using a phonemizer and an embedding layer. To synthesize audio from the output speech units, we utilize unit-based HiFi-GAN vocoders following \cite{lee2021textless,kong2020hifi}.

%------------------------------------ Table 1
%#############################################
\begin{table}[t]
\renewcommand{\arraystretch}{1.2}
\renewcommand{\tabcolsep}{1mm}
\centering
\caption{Training dataset statistics for each language pair (hours of source speech, `VoxPopuli (+ mTEDx)').}
\label{table:1}
\resizebox{0.999\linewidth}{!}{
\begin{tabular}{cccccccccccccccc}
\Xhline{3\arrayrulewidth}
{$\mathbf{L_s}\,$\textbackslash$\,\mathbf{L_t}$} & \textbf{Cs} & \textbf{Da} & \textbf{De} & \textbf{En} & \textbf{Es} & \textbf{Fi} & \textbf{Fr} & \textbf{Hu} & \textbf{It} & \textbf{Lt} & \textbf{Nl} & \textbf{Pl} & \textbf{Ro} & \textbf{Sk} & \textbf{Sl} \\ \hline

\textbf{Cs} &   - &  13 &  24 &  28 &  20 &  20 &  20 &  16 &  22 &  17 &  15 &  25 &  21 &  40 &  20 \\
\textbf{De} & 135 & 116 &   - & 120 & 126 & 136 & 124 & 127 & 136 &  99 & 124 & 133 & 127 & 134 & 135 \\
\textbf{El} &   - &   - &   - & - (+24) &  - &   - &   - &   - &   - &   - &   - &   - &   - &   - &   - \\
\textbf{En} & 280 & 250 & 302 &   - & 285 & 290 & 282 & 245 & 305 & 258 & 260 & 283 & 298 & 287 & 280 \\
\textbf{Es} &  81 &  59 &  94 & 91 (+176) & - & 84 &  94 &  62 & 104 &  56 &  77 &  78 &  90 &  78 &  72 \\
\textbf{Fi} &   7 &   5 &   9 &   9 &   7 &   - &   8 &   6 &   7 &   6 &   6 &   7 &   7 &   7 &   6 \\
\textbf{Fr} &  95 &  74 & 104 & 96 (+175) & 105 & 94 & - &  77 & 114 &  79 &  87 &  95 & 110 &  96 &  86 \\
\textbf{Hr} &  21 &  17 &  23 &  28 &  21 &  21 &  24 &  18 &  24 &  18 &  19 &  23 &  21 &  23 &  33 \\
\textbf{Hu} &  14 &   9 &  20 &  16 &  13 &  16 &  13 &   - &  15 &  11 &  10 &  15 &  15 &  15 &  13 \\
\textbf{It} &  63 &  57 &  72 & 67 (+99) & 76 & 65 &  73 &  56 &   - &  47 &  61 &  66 &  71 &  62 &  64 \\
\textbf{Lt} &   1 &   0 &   1 &   1 &   1 &   1 &   1 &   1 &   1 &   - &   1 &   1 &   1 &   1 &   1 \\
\textbf{Nl} &  14 &  14 &  25 &  18 &  16 &  19 &  20 &  14 &  21 &  10 &   - &  15 &  13 &  15 &  12 \\
\textbf{Pl} &  41 &  20 &  38 &  39 &  31 &  31 &  31 &  24 &  39 &  29 &  24 &   - &  31 &  37 &  33 \\
\textbf{Pt} &   - &   - &   - & - (+151) & - &   - &   - &   - &   - &   - &   - &   - &   - &   - &   - \\
\textbf{Ro} &  28 &  16 &  33 &  34 &  31 &  27 &  32 &  21 &  33 &  21 &  23 &  26 &   - &  28 &  25 \\
\textbf{Ru} &   - &   - &   - & - (+49) &  - &   - &   - &   - &   - &   - &   - &   - &   - &   - &   - \\
\textbf{Sk} &  15 &   5 &  10 &  10 &   7 &   7 &   6 &   6 &   8 &   6 &   6 &   9 &   7 &   - &   7 \\
\textbf{Sl} &   5 &   3 &   6 &   5 &   4 &   4 &   4 &   4 &   5 &   4 &   3 &   4 &   5 &   5 &   - \\

\Xhline{3\arrayrulewidth}
\end{tabular}}
\end{table}

%#############################################
%------------------------------------ Table 2
%#############################################
\begin{table}[t]
\renewcommand{\arraystretch}{1.3}
\renewcommand{\tabcolsep}{3mm}
\centering
\caption{Off-the-shelf ASR models for each language.}
\label{table:2}
\resizebox{0.9\linewidth}{!}{
\begin{tabular}{cc}
\Xhline{3\arrayrulewidth}
\textbf{Language} & \textbf{HuggingFace Model Name} \\ \hline
En & facebook/wav2vec2-large-960h-lv60-self \\
Es & jonatasgrosman/wav2vec2-large-xlsr-53-spanish \\
Fr & jonatasgrosman/wav2vec2-large-fr-voxpopuli-french \\
It & jonatasgrosman/wav2vec2-large-xlsr-53-italian \\
De & jonatasgrosman/wav2vec2-xls-r-1b-german \\
Nl & jonatasgrosman/wav2vec2-xls-r-1b-dutch \\
\Xhline{3\arrayrulewidth}
\end{tabular}}
\end{table}
%#############################################

\section{Experimental Setup}
\subsection{Multilingual Dataset}
\label{sec:4.1}
We train our model on speech-to-speech translation data containing 19 languages. We utilize VoxPopuli \cite{wang2021voxpopuli} and mTEDx \cite{salesky2021mtedx}. VoxPopuli is collected from European Parliament event recordings. We only utilize the subset of VoxPopuli, the aligned speech-to-speech data from the year 2013. It contains 15 source languages: English (En), German (De), French (Fr), Spanish (Es), Polish (Pl), Italian (It), Romanian (Ro), Hungarian (Hu), Czech (Cs), Dutch (Nl), Finnish (Fi), Croatian (Hr), Slovak (Sk), Slovene (Sl), and Lithuanian (Lt) and their human oral interpretation into 15 target languages: En, De, Fr, Es, Pl, It, Ro, Hu, Cs, Nl, Fi, Sk, Sl, Lt, and Danish (Da). The total duration of this VoxPopuli dataset is 11.2k hours. Since Hr exists only on the source side and Da only on the target side, a total of 16 languages are included in the VoxPopuli speech-to-speech dataset. In addition to VoxPopuli, we use mTEDx to increase the diversity of language. mTEDx is collected from the TEDx program and it provides speech-to-text translation data. We employ a subset of the translated dataset of \cite{anwar2023muavic}, which contains the translation of Es, Fr, Portuguese (Pt), It, Russian (Ru), and Greek (El) languages to English. Since mTEDx does not include target speech in audio, we generate the target speech by using a pre-trained T2S model \cite{kim2021conditional} on a single speaker dataset, LJSpeech \cite{ljspeech17}. The total duration of the used dataset from mTEDx is 0.7k hours. By combining the two datasets, we form about 12k hours of speech-to-speech translation data containing 19 languages. The detailed hours of each language pair can be found in Table \ref{table:1}.

\subsection{Evaluation Metric}
For the translation tasks, S2ST and T2ST, we measure BLEU \cite{post-2018-call} and COMET \cite{freitag2022comet}. The BLEU score is calculated by comparing the ground-truth transcription to a prediction generated by transcribing the synthesized audio using off-the-shelf ASR models available in HuggingFace\footnote{https://huggingface.co/models}. Following \cite{lee2021textless,duquenne2022speechmatrix}, we use the same pre-trained models with them as shown in Table \ref{table:2}. Similar to BLEU, COMET is also measured using the transcribed text. We use publicly available `wmt22-comet-da' model\footnote{https://github.com/Unbabel/COMET}. In general, the higher BLEU and COMET scores represent the better quality of the translation system. For evaluating the multilingual T2S system, we perform a human study and report the Mean Opinion Score (MOS).

\subsection{Implementation Details}
The UTUT model is composed of an encoder embedding layer, 12 Transformer encoder layers, 12 Transformer decoder layers, and decoder embedding layers. The unit vocabulary size is 1,000 and the embedding dimension is 1,024. For both the unit encoder and unit decoder, 8 attention heads, and a feed-forward dimension of 4,096 are utilized. The model is initialized from unit-based mBART \cite{popuri2022enhanced} trained on En and Es data. When we employ a pre-trained speech encoder, the unit embedding layer in the encoder is omitted and a convolutional adapter is inserted between the pre-trained speech encoder and the unit encoder to mitigate the length differences. The convolutional adapter is composed of 1 layer 1-dimensional convolution with stride 2 and a kernel size of 5 to downsample the audio features by a factor of 2. When we employ text inputs instead of speech, the unit embedding layer in the encoder is re-initialized and trained to embed the text inputs (\ie, phoneme tokens). For the unit-based HiFi-GAN vocoder, we directly utilized the pre-trained models of \cite{lee2021textless} for En, Es, and Fr. For It, De, and Nl, we train them for 1M steps on each language dataset with 1 GPU and 16 batch size. M-AILABS dataset for It, CSS10 \cite{park2019css10} dataset for De and Nl are employed.

\section{Speech-to-Speech Translation}
In this section, we show the effectiveness of the proposed UTUT on S2ST, especially in textless many-to-many spoken language translation using a single trained model.
\subsection{Experimental Settings}
\subsubsection{Evaluation Dataset}
We evaluate S2ST performances on a popular benchmark database, Europarl-ST \cite{iranzo2020europarl}. Europarl-ST is constructed using the debates carried out in European Parliament from the year 2012 and before so it has the same data domain as the VoxPopuli. We evaluate S2ST on all translation directions from (En, Es, Fr, It, De, Nl, Pl, Ro, Pt) to (En, Es, Fr, It, De, Nl) except the same language directions, using a single model. 

We also evaluate S2ST performances on CVSS-C \cite{jia2022cvss}. CVSS-C is a public S2ST dataset based on CoVoST2 \cite{wang2021covost}. It provides 21 languages to En translation. The 21 languages are Arabic (Ar), Catalan (Ca), Welsh (Cy), De, Estonian (Et), Es, Persian (Fa), Fr, Indonesian (Id), It, Japanese (Ja), Latvian (Lv), Mongolian (Mn), Nl, Pt, Ru, Sl, Swedish (Sv), Tamil (Ta), Turkish (Tr), and Chinese (Zh). Since 13 languages out of 21 are novel from the pre-training dataset, we can evaluate whether UTUT can be easily adapted to new languages.

\subsubsection{Finetuning}
We explore two kinds of our model; The first is without finetuning and the second is with finetuning by attaching a speech encoder, XLS-R 1B \cite{babu2021xls} finetuned on CoVoST-2 \cite{wang2021covost}. When we directly evaluate the trained UTUT model on Europarl-ST without finetuning, input speech is quantized into speech units with the same speech unit extraction process as done in training. When we attach XLS-R instead of using the quantizer, we finetune the UTUT model 50k steps on the same dataset used for training by freezing the pre-trained speech encoder XLS-R. For CVSS-C dataset, the trained model is finetuned for 50k steps on CVSS-C dataset by using a quantizer.

%------------------------------------ Table 3
%#############################################
\begin{table*}[t]
\renewcommand{\arraystretch}{1.4}
\renewcommand{\tabcolsep}{3.5mm}
\centering
\caption{Multilingual Speech-to-Speech Translation (S2ST) results (BLEU) on Europarl-ST. $^\dagger$The results of Unidirectional S2ST \cite{duquenne2022speechmatrix} are reported by using 48 unidirectional translation models for each direction. Shaded results are for new language pairs that do not exist as pairs in the training data.}
\label{table:3}
\resizebox{0.999\linewidth}{!}{
\begin{tabular}{cccccccc}
\Xhline{3\arrayrulewidth}
\multirow{2}{*}{$\mathbf{L_s}\,$\textbackslash$\,\mathbf{L_t}$} & \multicolumn{6}{c}{\makecell{\textbf{BLEU} \\ Unidirectional S2ST$^\dagger$ \cite{duquenne2022speechmatrix} \, $|$ \, Enhanced S2ST \cite{popuri2022enhanced} \, $|$ \, \textbf{UTUT}}} \\ \cline{2-8}
& \textbf{En} & \textbf{Es} & \textbf{Fr} & \textbf{It} & \textbf{De} & \textbf{Nl} \\ \hline
\textbf{En} &         -          & 21.9 $|$ 21.7 $|$ 23.4 & 19.2 $|$ 20.5 $|$ 21.2 &  11.5 $|$ 12.9 $|$ 11.9 & 10.1 $|$ \;\:9.4 $|$ \;\:9.8 & 15.1 $|$ 11.0 $|$ 12.2 \\

\textbf{Es} & 20.4 $|$ 24.2 $|$ 21.3 &         -          & 16.3 $|$ 14.8 $|$ 18.9 & 11.1 $|$ \;\:9.7 $|$ 12.4 & \;\:6.1 $|$ \;\:4.5 $|$ \;\:5.3 &   \;\:8.0 $|$ \;\:7.0 $|$ \;\:8.0 \\

\textbf{Fr} & 20.7 $|$ 23.0 $|$ 20.7 &  18.4 $|$ 16.1 $|$ 19.8 &         -          & 10.2 $|$ 10.8 $|$ 12.6 & \;\:6.3 $|$ \;\:5.4 $|$ \;\:5.7  & \;\:8.4 $|$ \;\:7.1 $|$ \;\:8.7 \\

\textbf{It} & 18.9 $|$ 23.1 $|$ 19.7 & 19.6 $|$ 15.3 $|$ 19.4 & 15.3 $|$ 14.0 $|$ 15.0 &         -          & \;\:4.9 $|$ \;\:4.5 $|$ \;\:5.1  &  \;\:6.5 $|$ \;\:6.7 $|$ \;\:7.5 \\

\textbf{De} & 16.3 $|$ 17.6 $|$ 15.8 & 11.7 $|$ 13.7 $|$ 13.1 & 10.7 $|$ 13.6 $|$ 13.2 & \;\:3.8 $|$ \;\:7.2 $|$ \;\:6.3 &         -          & 10.4 $|$ \;\:9.7 $|$ 10.8  \\

\textbf{Nl} & 18.0 $|$ 19.7 $|$ 16.9 & 13.2 $|$ 12.8 $|$ 12.3 & 13.0 $|$ 12.6 $|$ 12.8 & \;\:5.2 $|$ \;\:7.4 $|$ \;\:6.6 & \;\:8.1 $|$ \;\:6.3 $|$ \;\:6.6 &         -          \\

\textbf{Pl} &\;\:4.9 $|$ 19.1 $|$ 17.2 & \;\:6.3 $|$ 13.2 $|$ 14.1 & \;\:5.5 $|$ 12.4 $|$ 13.8 & \;\:5.8 $|$ \;\:6.9 $|$ \;\:7.1 & \;\:2.8 $|$ \;\:5.7 $|$ \;\:5.6 & \;\:1.6 $|$ \;\:7.5 $|$ \;\:8.0 \\

\textbf{Ro} &  22.6 $|$ 24.3 $|$ 21.9 &  20.1 $|$ 17.2 $|$ 19.9 & 18.6 $|$ 17.7 $|$ 19.5 & \;\:8.7 $|$ 10.7 $|$ 11.0 & \;\:6.5 $|$ \;\:5.9 $|$ \;\:6.0 & \;\:3.5 $|$ \;\:8.4 $|$ \;\:9.2 \\

\textbf{Pt} & 21.2 $|$ 25.1 $|$ 21.5 & 23.2 $|$ \colorbox{gray!20}{\makebox[1.0em]{15.8}} $|$ \colorbox{gray!20}{\makebox[1.0em]{19.5}} & 18.1 $|$ \colorbox{gray!20}{\makebox[1.0em]{14.5}} $|$ \colorbox{gray!20}{\makebox[1.0em]{16.9}}  & \;\:4.4 $|$ \colorbox{gray!20}{\makebox[1.0em]{9.7}} $|$ \colorbox{gray!20}{\makebox[1.0em]{11.4}} & \;\:4.7 $|$ \colorbox{gray!20}{\makebox[1.0em]{\;\:4.2}} $|$ \colorbox{gray!20}{\makebox[1.0em]{\;\:5.3}} &  \;\:5.0 $|$ \colorbox{gray!20}{\makebox[1.0em]{\;\:6.7}} $|$ \colorbox{gray!20}{\makebox[1.0em]{\;\:8.4}} \\

\Xhline{3\arrayrulewidth}
\end{tabular}}
\end{table*}

%#############################################

%------------------------------------ Table 4
%#############################################
\begin{table*}[t]
\renewcommand{\arraystretch}{1.4}
\renewcommand{\tabcolsep}{2.mm}
\centering
\caption{Multilingual Speech-to-Speech Translation (S2ST) results (COMET) on Europarl-ST. $^\dagger$The results of Unidirectional S2ST \cite{duquenne2022speechmatrix} are reported by using 48 unidirectional translation models for each direction. Shaded results are for new language pairs that do not exist as pairs in the training data.}
\label{table:3comet}
\resizebox{0.999\linewidth}{!}{
\begin{tabular}{cccccccc}
\Xhline{3\arrayrulewidth}
\multirow{2}{*}{$\mathbf{L_s}\,$\textbackslash$\,\mathbf{L_t}$} & \multicolumn{6}{c}{\makecell{\textbf{COMET} \\ Unidirectional S2ST$^\dagger$ \cite{duquenne2022speechmatrix} \, $|$ \, Enhanced S2ST \cite{popuri2022enhanced} \, $|$ \, \textbf{UTUT}}} \\ \cline{2-8}
& \textbf{En} & \textbf{Es} & \textbf{Fr} & \textbf{It} & \textbf{De} & \textbf{Nl} \\ \hline

\textbf{En} &         -          & 0.647 $|$ 0.661 $|$ 0.701 & 0.646 $|$ 0.637 $|$ 0.689 & 0.549 $|$ 0.587 $|$ 0.599 & 0.526 $|$ 0.517 $|$ 0.567 & 0.624 $|$ 0.526 $|$ 0.585 \\

\textbf{Es} & 0.676 $|$ 0.730 $|$ 0.709 & - & 0.598 $|$ 0.541 $|$ 0.649 & 0.577 $|$ 0.526 $|$ 0.605 & 0.498 $|$ 0.413 $|$ 0.494 & 0.511 $|$ 0.435 $|$ 0.519 \\

\textbf{Fr} & 0.684 $|$ 0.718 $|$ 0.707 & 0.626 $|$ 0.608 $|$ 0.677 & - & 0.577 $|$ 0.564 $|$ 0.605 & 0.492 $|$ 0.443 $|$ 0.501 & 0.523 $|$ 0.458 $|$ 0.529 \\

\textbf{It} & 0.665 $|$ 0.718 $|$ 0.697 & 0.646 $|$ 0.592 $|$ 0.681 & 0.600 $|$ 0.552 $|$ 0.637 & - & 0.465 $|$ 0.422 $|$ 0.487 & 0.489 $|$ 0.446 $|$ 0.504 \\

\textbf{De} & 0.659 $|$ 0.680 $|$ 0.676 & 0.574 $|$ 0.617 $|$ 0.624 & 0.556 $|$ 0.578 $|$ 0.609 & 0.456 $|$ 0.531 $|$ 0.533 & - & 0.563 $|$ 0.511 $|$ 0.564 \\

\textbf{Nl} & 0.649 $|$ 0.687 $|$ 0.673 & 0.583 $|$ 0.593 $|$ 0.607 & 0.561 $|$ 0.559 $|$ 0.597 & 0.452 $|$ 0.507 $|$ 0.521 & 0.529 $|$ 0.461 $|$ 0.517 & - \\

\textbf{Pl} & 0.660 $|$ 0.682 $|$ 0.676 & 0.602 $|$ 0.586 $|$ 0.639 & 0.572 $|$ 0.547 $|$ 0.611 & 0.480 $|$ 0.515 $|$ 0.542 & 0.507 $|$ 0.457 $|$ 0.519 & 0.457 $|$ 0.468 $|$ 0.526 \\

\textbf{Ro} & 0.709 $|$ 0.749 $|$ 0.724 & 0.640 $|$ 0.626 $|$ 0.686 & 0.640 $|$ 0.604 $|$ 0.664 & 0.547 $|$ 0.575 $|$ 0.610 & 0.517 $|$ 0.471 $|$ 0.528 & 0.467 $|$ 0.488 $|$ 0.549 \\

\textbf{Pt} & 0.691 $|$ 0.727 $|$ 0.683 & 0.693 $|$ \colorbox{gray!20}{\makebox[1.45em]{0.578}} $|$ \colorbox{gray!20}{\makebox[1.45em]{0.641}} & 0.637 $|$ \colorbox{gray!20}{\makebox[1.45em]{0.548}} $|$ \colorbox{gray!20}{\makebox[1.45em]{0.588}} & 0.577 $|$ \colorbox{gray!20}{\makebox[1.45em]{0.530}} $|$ \colorbox{gray!20}{\makebox[1.45em]{0.576}} & 0.482 $|$ \colorbox{gray!20}{\makebox[1.45em]{0.419}} $|$ \colorbox{gray!20}{\makebox[1.45em]{0.480}} & 0.488 $|$ \colorbox{gray!20}{\makebox[1.45em]{0.444}} $|$ \colorbox{gray!20}{\makebox[1.45em]{0.499}} \\

\Xhline{3\arrayrulewidth}
\end{tabular}}
\end{table*}

%#############################################

%------------------------------------ Table 5
%#############################################
\begin{table*}[t]
\renewcommand{\arraystretch}{1.4}
\renewcommand{\tabcolsep}{3.5mm}
\centering
\caption{Comparison of dataset statistics, specifically the number of hours of source speech, used for training UTUT and Unidirectional S2ST \cite{duquenne2022speechmatrix} for each language pair. Please note that UTUT is trained with many-to-many language S2ST while Unidirectional S2ST is trained on only the target language pair data.}
\label{table:4}
\resizebox{0.75\linewidth}{!}{
\begin{tabular}{cccccccc}
\Xhline{3\arrayrulewidth}
\multicolumn{7}{c}{Training Data Statistics: Hours of Paired Language (Unidirectional S2ST $|$ UTUT)} \\ \hline
{$\mathbf{L_s}\,$\textbackslash$\,\mathbf{L_t}$} & \textbf{En} & \textbf{Es} & \textbf{Fr} & \textbf{It} & \textbf{De} & \textbf{Nl} \\ \hline
\textbf{En} & - & 4715 $|$ 552 & 5178 $|$ 553 & 4897 $|$ 471 & 4676 $|$ 422 & 4422 $|$ 278 \\
\textbf{Es} & 4708 $|$ 552 & - & 4446 $|$ 199 & 4418 $|$ 180 & 3041 $|$ 220 & 3067 $|$ \;\:93 \\
\textbf{Fr} & 5171 $|$ 553 & 4455 $|$ 199 & - & 4618 $|$ 187 & 3457 $|$ 228 & 3273 $|$ 107 \\
\textbf{It} & 4948 $|$ 471 & 4500 $|$ 180 & 4700 $|$ 187 & - & 3460 $|$ 208 & 3414 $|$ \;\:82 \\
\textbf{De} & 4734 $|$ 422 & 3113 $|$ 220 & 3536 $|$ 228 & 3476 $|$ 208 & - & 3384 $|$ 149 \\
\textbf{Nl} & 4396 $|$ 278 & 3066 $|$ \;\:93 & 3269 $|$ 107 & 3355 $|$ \;\:82 & 3305 $|$ 149 & - \\
\textbf{Pl} & 3662 $|$ 322 & 2735 $|$ 109 & 2913 $|$ 126 & 2883 $|$ 105 & 2646 $|$ 171 & 2540 $|$ \;\:39 \\
\textbf{Ro} & 2290 $|$ 332 & 1894 $|$ 121 & 2024 $|$ 142 & 1996 $|$ 104 & 1275 $|$ 160 & 1384 $|$ \;\:36 \\
\textbf{Pt} & 3606 $|$ 151 & 3525 $|$ \makebox[1.5em]{\;\:-\;\:} & 3421 $|$ \makebox[1.5em]{\;\:-\;\:} & 3403 $|$ \makebox[1.5em]{\;\:-\;\:} & 2224 $|$ \makebox[1.5em]{\;\:-\;\:} & 2436 $|$ \makebox[1.5em]{\;\:-\;\:} \\
\Xhline{3\arrayrulewidth}
\end{tabular}}
\end{table*}

%#############################################

\subsection{Experimental Results}
\subsubsection{Textless Many-to-Many S2ST}
Table \ref{table:3} shows the S2ST results on Europarl-ST. Since no baseline method can perform textless many-to-many spoken languages S2ST, we compare the performances with multiple state-of-the-art unidirectional S2ST models \cite{duquenne2022speechmatrix} which are the textless version of \cite{lee2021directs2s}. In addition, for comparison, we also fine-tune an Es-En pre-trained model from Enhanced S2ST \cite{popuri2022enhanced} on the same many-to-many language data used by UTUT. Therefore, the performances of Enhanced S2ST and UTUT are obtained using a single many-to-many language S2ST model while \cite{duquenne2022speechmatrix} uses 48 models to handle multiple unidirectional translation tasks. Overall, the performance of UTUT shows better or comparable results on 32 language pairs out of 48 compared to the unidirectional translation models, with much smaller training data. To compare the amount of training data, we show the amount of data of each method for each language pair in Table \ref{table:4}. By comparing with the unidirectional S2ST model, the UTUT is trained on much smaller data. For example, no data is utilized for Pt-X except for Pt-En, and 21 times less data is utilized for overall X-It translation. This is possible since the proposed method is trained in a many-to-many translation setting so that the model can learn not only the target language pair but also from other language pairs  (Please refer to Sec. \ref{sec:3.3.2}). This result indicates that we can greatly reduce training costs and save memories by using a single UTUT model instead of using multiple unidirectional S2ST models. Moreover, compared to the Enhanced S2ST \cite{popuri2022enhanced}, the proposed UTUT achieves better overall performance, except in the X-to-En direction. This is because the Enhanced S2ST is fine-tuned from an Es-En translation model, and we found that fine-tuning a pre-trained unidirectional model to produce different languages is suboptimal, especially when there is insufficient target language data (\eg, X-to-Nl translation direction). Another important aspect to highlight is training efficiency. The Enhanced S2ST, which utilizes continuous speech as inputs, requires significantly more training time and larger computing memory compared to the proposed UTUT, which uses discretized speech units for both inputs and outputs \cite{chang2023exploring}. In Table \ref{table:3comet}, the COMET performances of different S2ST systems are shown. Similar to the BLEU results, the proposed UTUT method achieves better performances overall in COMET. The results confirm the effectiveness of the proposed unit-to-unit translation training method in many-to-many language translation setting in both data efficiency and performances.

%------------------------------------ Table 6
%#############################################
\begin{table}[t]
\renewcommand{\arraystretch}{1.3}
\renewcommand{\tabcolsep}{3.3mm}
\centering
\caption{Multilingual Speech-to-Speech Translation (S2ST) results (BLEU) using a pre-trained speech encoder on Europarl-ST. Shaded results are for new language pairs that do not exist as pairs in the training data.}
\label{table:3_xlsr}
\resizebox{0.999\linewidth}{!}{
\begin{tabular}{ccccccc}
\Xhline{3\arrayrulewidth}
\multirow{2}{*}{$\mathbf{L_s}\,$\textbackslash$\,\mathbf{L_t}$} & \multicolumn{6}{c}{\textbf{BLEU}} \\ \cline{2-7}
& \textbf{En} & \textbf{Es} & \textbf{Fr} & \textbf{It} & \textbf{De} & \textbf{Nl} \\ \hline

\textbf{En} & - & 22.4 & 21.0 & 12.2 & 8.8 & 10.6 \\
\textbf{Es} & 21.9 & - & 18.4 & 12.1 & 4.6 & 7.2 \\
\textbf{Fr} & 21.6 & 19.7 & - & 12.9 & 5.1 & 7.5 \\
\textbf{It} & 20.2 & 20.3 & 17.3 & - & 4.7 & 6.8 \\
\textbf{De} & 17.4 & 13.6 & 14.2 & 6.5 & - & 9.9 \\
\textbf{Nl} & 19.1 & 13.4 & 13.9 & 6.8 & 6.6 & - \\
\textbf{Pl} & 18.6 & 14.5 & 14.1 & 7.1 & 5.2 & 7.4 \\
\textbf{Ro} & 23.2 & 19.9 & 20.1 & 11.0 & 5.8 & 8.5 \\
\textbf{Pt} & 26.5 & \colorbox{gray!20}{23.9} & \colorbox{gray!20}{20.1}  & \colorbox{gray!20}{13.8} & \colorbox{gray!20}{5.9} & \colorbox{gray!20}{9.4} \\

\Xhline{3\arrayrulewidth}
\end{tabular}}
\end{table}

%#############################################

By replacing the quantizer with a multilingual speech encoder, XLS-R, we can enhance performance, particularly for source languages with few training samples, as shown in Table \ref{table:3_xlsr}. he results indicate the performance of the UTUT pre-trained model can be improved by incorporating a large-scale trained speech encoder. 

Finally, the shaded region in Table \ref{table:3}, Table \ref{table:3comet}, and Table \ref{table:3_xlsr} represents the results for language pairs that do not exist in the training data as pairs, which means the model successfully learned to translate Portuguese (Pt) to other languages by using only Pt-to-En pair data. The results show that UTUT can translate the speech between the novel language pairs where the paired data does not exist, by learning to comprehend the source language in the unit encoder and to express the comprehended semantics into the target language in the unit decoder. The samples can be found on our demo page\footnote{\url{https://choijeongsoo.github.io/utut}}.

\subsubsection{Many-to-English S2ST}
Table \ref{table:5} shows an additional S2ST result on CVSS-C \cite{jia2022cvss} dataset. For CVSS-C, we mainly compare our method with the previous methods that do not use large-scale multilingual pre-trained speech encoders which are jointly finetuned during S2ST training since they are not publicly available (\ie, w2v-BERT \cite{chung2021w2v} and mSLAM \cite{bapna2022mslam}). By comparing with the previous methods, Translatotron \cite{jia2019translatotron} and Translatotron2 \cite{jia2021translatotron2}, we confirm that the proposed UTUT model exhibits strong performance even trained without the use of text supervision, in contrast to the previous methods that utilize text supervision. Moreover, we can find that the performances for the mid- and low-resource languages are much higher than the previous methods. The results confirm that the proposed many-to-many language translation training strategy is more effective for the low-resource languages. Please note that developing textless S2ST systems is important because there are languages without writing systems \cite{lee2021textless} (\eg, dialects), for which text-based systems cannot be developed. Finally, the proposed model also has the advantage compared to the ASR-NMT-T2S cascaded system. As the quantizer and the vocoder are fully trainable in a self-supervised manner, it is easier to scale the model to different languages and large datasets, while ASR and T2S require the labeled dataset to be trained. 

%------------------------------------ Table 7
%#############################################
\begin{table*}[t]
\renewcommand{\arraystretch}{1.4}
\renewcommand{\tabcolsep}{1.3mm}
\centering
\caption{Performance (BLEU) comparisons of S2ST from 21 languages to En on CVSS-C.\protect\footnotemark}
\label{table:5}
\resizebox{0.9999\linewidth}{!}{
\begin{tabular}{l c cccc ccccc cccccccccccc}
\Xhline{3\arrayrulewidth}
\multirow{3}{*}{\textbf{Method}} & \multirow{3}{*}{\textbf{Avg.}} & \multicolumn{4}{c}{\textbf{High}} & \multicolumn{5}{c}{\textbf{Mid}} & \multicolumn{12}{c}{\textbf{Low}} \\   \cmidrule(l{2pt}r{2pt}){3-6}\cmidrule(l{2pt}r{2pt}){7-11}\cmidrule(l{2pt}r{2pt}){12-23}
& & Fr & De & Ca & Es & Fa & It & Ru & Zh & Pt & Nl & Tr & Et & Mn & Ar & Lv & Sl & Sv & Cy & Ta & Ja & Id \\ 
& & 264 & 184 & 136 & 113 & 49 & 44 & 18 & 10 & 10 & 7 & 4.1 & 3.4 & 3.0 & 2.1 & 2.1 & 2.0 & 1.7 & 1.7 & 1.6 & 1.3 & 1.2\\
\cmidrule(l{2pt}r{2pt}){1-1}\cmidrule(l{2pt}r{2pt}){2-2}\cmidrule(l{2pt}r{2pt}){3-6}\cmidrule(l{2pt}r{2pt}){7-11}\cmidrule(l{2pt}r{2pt}){12-23}
$\bullet$\textbf{\textit{ Cascaded system}} \\
\,\,\, S2TT + T2S \cite{jia2021png,jia2022cvss} & 10.6 & 31.2 & 23.9 & 26.8 & 33.3 & 3.4 & 28.1 & 24.4 & 6.8 & 14.8 & 9.8 & 5.1 & 1.7 & 0.3 & 4.1 & 2.3 & 0.6 & 1.4 & 2.1 & 0.2 & 0.7 & 0.9 \\ \cmidrule(l{2pt}r{2pt}){1-1}\cmidrule(l{2pt}r{2pt}){2-2}\cmidrule(l{2pt}r{2pt}){3-6}\cmidrule(l{2pt}r{2pt}){7-11}\cmidrule(l{2pt}r{2pt}){12-23}
$\bullet$\textbf{\textit{ S2ST system}} \\
\,\,\, Translatotron \cite{jia2019translatotron} & 3.4 & 15.5 & 6.9 & 11.0 & 14.1 & 1.4 & 9.3 & 4.3 & 1.5 & 2.2 & 2.1 & 1.2 & 0.1 & 0.1 & 0.1 & 0.2 & 0.3 & 0.4 & 0.3 & \textbf{0.1} & 0.2 & 0.1 \\
\,\,\, Translatotron2 \cite{jia2021translatotron2,jia2022cvss} & 8.7 & \textbf{28.3} & \textbf{19.7} & \textbf{23.5} & \textbf{30.1} & 2.4 & \textbf{24.1} & 19.6 & \textbf{4.5} & 12.5 & 6.5 & \textbf{3.8} & 0.6 & 0.2 & \textbf{1.7} & 1.5 & 0.4 & 1.3 & \textbf{0.9} & \textbf{0.1} & \textbf{0.5} & 0.4 \\
\,\,\, \textbf{UTUT} & \textbf{11.2} & 26.9 & 19.0 & 22.9 & 27.7 & \textbf{4.2} & 23.4 & \textbf{26.6} & 2.1 & \textbf{25.6} & \textbf{16.5} & 3.2 & \textbf{3.6} & \textbf{0.3} & 1.1 & \textbf{6.3} & \textbf{14.7} & \textbf{8.8} & 0.6 & \textbf{0.1} & 0.4 & \textbf{0.6}\\ \hline

\multicolumn{14}{l}{$\bullet$\textbf{\textit{ S2ST system (using large-scale multilingual pre-trained speech encoder)}}} \\
\,\,\, Textless S2ST + w2v-BERT \textit{(Textless)} \cite{li2022textlesss2s} & 17.7 & - & - & - & - & - & - & - & - & - & - & - & - & - & - & - & - & - & - & - & - & - \\
\,\,\, Translatotron2 + w2v-BERT \cite{jia2021translatotron2,jia2022leveraging} & 17.9 & 33.6 & 30.6 & 30.1 & 35.9 & 6.0 & 32.5 & 38.9 & 5.2 & 31.9 & 29.3 & 9.2 & 16.0 & 0.2 & 10.4 & 15.6 & 17.8 & 25.9 & 4.2 & 0.3 & 0.9 & 1.5 \\
\,\,\, Translatotron2 + mSLAM \cite{jia2021translatotron2,jia2022leveraging} & 19.3 & 33.9 & 31.5 & 30.6 & 36.8 & 7.2 & 33.7 & 41.6 & 6.4 & 34.1 & 31.1 & 16.1 & 17.1 & 0.3 & 10.0 & 14.4 & 22.9 & 28.4 & 5.4 & 0.2 & 1.3 & 2.5 \\
\,\,\, UnitY + w2v-BERT \cite{inaguma2022unity} & 24.5 & 35.2 & 32.6 & 33.3 & 37.2 & 14.9 & 35.0 & 42.3 & 10.8 & 41.7 & 32.5 & 22.2 & 18.7 & 2.7 & 24.6 & 21.3 & 26.6 & 34.1 & 16.5 & 1.8 & 8.0 & 22.9 \\ 
\Xhline{3\arrayrulewidth}
\end{tabular}}
\end{table*}
%#############################################

%------------------------------------ Table 8
%#############################################
\begin{table}[t]
\renewcommand{\arraystretch}{1.2}
\renewcommand{\tabcolsep}{3.5mm}
\centering
\caption{MOS result comparisons of Multilingual Text-to-Speech Synthesis (T2S) on Europarl-ST.}
\label{table:6}
\resizebox{0.999 \linewidth}{!}{
\begin{tabular}{ccccccc}
\Xhline{3\arrayrulewidth}
\textbf{Method} & \textbf{En} & \textbf{Es} & \textbf{Fr} & \textbf{It} & \textbf{De} & \textbf{Nl}\\ \hline
YourTTS & 3.35 & - & 3.46 & - & - & - \\
LAML & 3.53 & 3.47 & 3.14 & 3.77 & 3.67 & 3.81 \\
\textbf{UTUT} & 4.48 & 4.49 & 4.45 & 4.45 & 4.26 & 4.50 \\
\Xhline{3\arrayrulewidth}
\end{tabular}}
\end{table}
%#############################################

\section{Multilingual Text-to-Speech Synthesis and Text-to-Speech Translation}
\footnotetext{The number below the language name represents hours of source speech in the train sets.}
In this section, we demonstrate the extensibility of the proposed UTUT system to text-related tasks, T2S and T2ST. By embedding phonemes transformed from multilingual text to the trained UTUT model, we can perform multilingual T2S with a single model by constructing text-to-unit translation. Moreover, by conditioning the unit decoder with the target language token, we can translate the input text into the desired speech in the target language, which is the T2ST task.

%------------------------------------ Table 9
%#############################################
\begin{table}[t]
\renewcommand{\arraystretch}{1.8}
\renewcommand{\tabcolsep}{3.0mm}
\centering
\caption{Multilingual Text-to-Speech Translation (T2ST) results on Europarl-ST. Shaded results are for novel language pairs not in the training data as pairs.}
\label{table:7}
\resizebox{0.999\linewidth}{!}{
\begin{tabular}{ccccccc}
\Xhline{3\arrayrulewidth}
\multirow{2}{*}{$\mathbf{L_s}\,$\textbackslash$\,\mathbf{L_t}$} & \multicolumn{6}{c}{\textbf{BLEU} / \textbf{COMET}} \\ \cline{2-7}
& \textbf{En} & \textbf{Es} & \textbf{Fr} & \textbf{It} & \textbf{De} & \textbf{Nl} \\ \hline

\textbf{En} & - & \makecell{23.4 \\ 0.712} & \makecell{23.2\\0.712} & \makecell{11.3\\0.594} & \makecell{10.0\\0.567} & \makecell{11.9\\0.585} \\ \hdashline
\textbf{Es} & \makecell{22.5\\0.712} & - & \makecell{19.3\\0.660} & \makecell{10.2\\0.586} & \makecell{4.7\\0.488} & \makecell{6.8\\0.507} \\  \hdashline
\textbf{Fr} & \makecell{22.5\\0.717} & \makecell{19.0\\0.676} & - & \makecell{11.4\\0.606} & \makecell{4.8\\0.495} & \makecell{7.5\\0.519} \\ \hdashline
\textbf{It} & \makecell{22.0\\0.709} & \makecell{19.8\\0.686} & \makecell{18.1\\0.653} & - & \makecell{4.7\\0.482} & \makecell{6.5\\0.494} \\ \hdashline
\textbf{De} & \makecell{17.3\\0.692} & \makecell{13.7\\0.643} & \makecell{14.7\\0.634} & \makecell{5.9\\0.536} & - & \makecell{10.7 \\0.565} \\ \hdashline
\textbf{Nl} & \makecell{18.1\\0.692} & \makecell{12.4\\0.632} & \makecell{14.6\\0.627} & \makecell{6.0\\0.509} & \makecell{6.7\\0.521} & - \\ \hdashline
\textbf{Pl} & \makecell{18.9\\0.701} & \makecell{14.1\\0.653} & \makecell{14.5\\0.634} & \makecell{6.9\\0.545} & \makecell{5.2\\0.504} & \makecell{7.0\\0.525} \\ \hdashline
\textbf{Ro} & \makecell{22.4\\0.730} & \makecell{18.8\\0.686} & \makecell{20.5\\0.685} & \makecell{10.0\\0.599} & \makecell{5.8\\0.509} & \makecell{8.5\\0.551} \\ \hdashline
\textbf{Pt} & \makecell{28.3\\0.755} & \makecell{\colorbox{gray!20}{22.5}\\  \colorbox{gray!20}{0.707}} & \makecell{\colorbox{gray!20}{20.2} \\ \colorbox{gray!20}{0.667}}  & \makecell{\colorbox{gray!20}{11.5}\\ \colorbox{gray!20}{0.607}} & \makecell{\colorbox{gray!20}{4.8}\\ \colorbox{gray!20}{0.485}} & \makecell{\colorbox{gray!20}{7.9} \\ \colorbox{gray!20}{0.517}} \\

\Xhline{3\arrayrulewidth}
\end{tabular}}
\end{table}

%#############################################

\subsection{Experimental Settings}
\subsubsection{Evaluation Dataset} 
We evaluate T2S performances on three benchmark databases, LRS3 \cite{afouras2018lrs3} for En, mTEDx \cite{salesky2021mtedx} for Es, Fr, and It, and Europarl-ST \cite{iranzo2020europarl} for En, Es, Fr, It, De, and Nl. For LRS3 and mTEDx, their test sets are used for the evaluation. For Europarl-ST, the source translation data from the direction of X-En are used while En-Es splits are used for En language generation. T2ST performances are measured on Europarl-ST \cite{iranzo2020europarl} for the directions from (En, Es, Fr, It, De, Nl, Pl, Ro, Pt) to (En, Es, Fr, It, De, Nl) similar to S2ST but using text inputs.

\subsubsection{Finetuning}
Firstly, we convert multilingual text into IPA phonemes using a phonemizer \cite{Bernard2021}. The resulting vocabulary size of phonemes is 83. Then, by setting the phonemes as inputs, we finetune the trained UTUT model by re-initializing only the embedding weight that was originally trained to embed the input speech units into features. We finetune the model for 100k steps on VoxPopuli and mTEDx datasets which are the same as that used for training, but with text-to-speech pair data that added the case of reconstruction. Then, the finetuned model is validated on the three databases, LRS3, mTEDx, and Europarl-ST.

\subsection{Experimental Results}
\subsubsection{Multilingual T2S}
Table \ref{table:6} shows the Mean Opinion Score (MOS) comparisons with state-of-the-art multilingual T2S systems, YourTTS \cite{casanova2022yourtts} and LAML \cite{lux2022low}. The MOS test is conducted by soliciting feedback from 15 participants who assess 10 audio samples for each of the 6 languages (En, Es, Fr, It, De, and Nl), assigning a score ranging from 1 to 5. Since YourTTS does not support Es, It, De, and Nl, only En and Fr languages are evaluated. The proposed system, UTUT, outperforms the previous methods in terms of speech synthesis performance. The generated samples from UTUT are more natural and achieve about a 4.44 mean score across the languages. Please note that even though the UTUT is pre-trained without using text, the model can easily transferred to operate with the text inputs as the speech units are easily associated with the phonetic information. Therefore, the proposed UTUT can achieve comparable performance with the previous multilingual T2S systems \cite{casanova2022yourtts,lux2022low} which are trained text-speech pair data from scratch, with a small finetuning steps on text-speech pair data.

\subsubsection{Many-to-Many T2ST}
Table \ref{table:7} shows T2ST performances (BLEU) obtained using a single UTUT model. By comparing the BLEU score of T2ST with the S2ST performance of UTUT (Table \ref{table:3}), we observe that the two translation systems exhibit similar levels of performance, so there is no drastic performance drop after changing into a text-based system. The results confirm that even though the UTUT model is not trained using text inputs, the model can be easily transferred into text-based systems. Furthermore, akin to S2ST, UTUT has the capability to perform translation between previously unseen language pairs, Pt-Es, Pt-Fr, Pt-It, Pt-De, and Pt-Nl (shaded region in the table).

%------------------------------------ Figure3
%#############################################
\begin{figure*}[t!]
\centering
\centerline{\includegraphics[width=18.1cm]{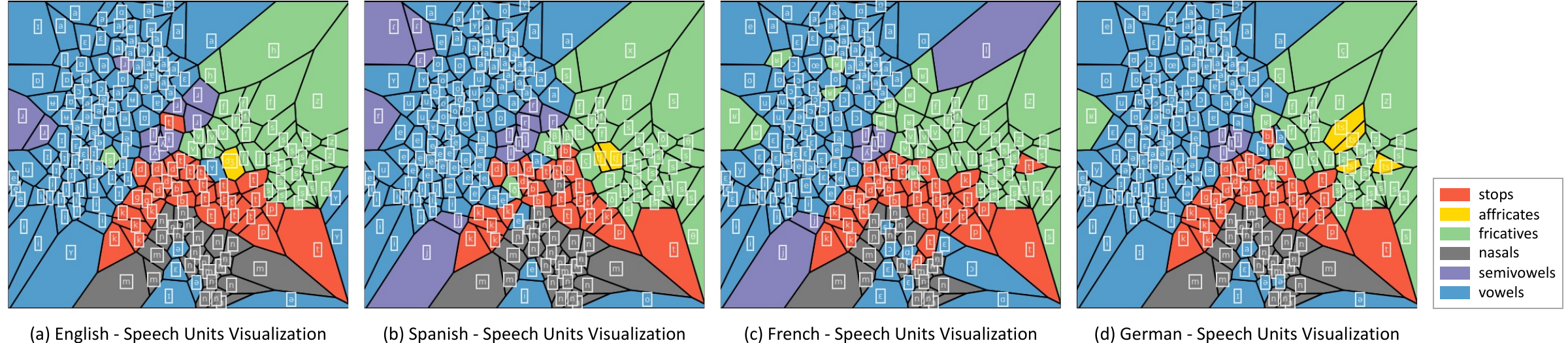}}
\vspace{-0.1cm}
\caption{Speech unit visualizations for En, Es, Fr, and De. For visualization, the top 200 frequently appearing units are employed.}
\label{fig:3}
\end{figure*}
%#############################################

%------------------------------------ Figure4
%#############################################
\begin{figure*}[t!]
\centering
\centerline{\includegraphics[width=18.1cm]{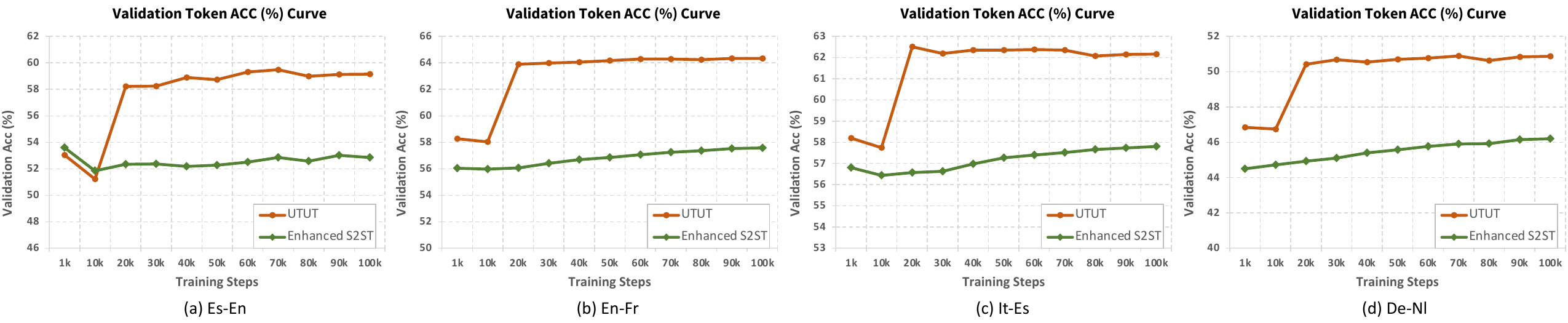}}
\vspace{-0.2cm}
\caption{Learning curves of the UTUT and the continuous speech-based S2ST method \cite{popuri2022enhanced} when transferred to T2ST. Token-level accuracy on the validation set is measured for each language translation direction: (a) Es-En, (b) En-Fr, (c) It-Es, and (d) De-Nl.}
\label{fig:4}
\end{figure*}
%#############################################

\section{Ablation Study}
\subsection{Transferability of the UTUT model to text inputs}
In order to analyze that how the trained UTUT model can be easily transferred to text-based systems, we visualize the speech unit representations following \cite{sicherman2023analysing}. To this end, we select 200 speech units from the 1,000 that appear most frequently in English (En), Spanish (Es), French (Fr), and German (De). The visualization results are shown in Fig. \ref{fig:3}. First, we can see that each speech unit can be mapped to different phonemes, indicating that the speech units contain linguistic information. Consequently, the visualization supports the notion that the association between speech units and phonemized text can be easily established. Second, we observe that similar phoneme families (\ie, indicated by color) are represented by the speech units across different languages, suggesting that the same set of speech units can be employed for multiple languages. Therefore, through visualization, we can confirm that using the same set of speech units to represent different languages is valid, and each speech unit contains phonetic information, facilitating an easy transfer to text input.

We also compare the convergence speed between the proposed UTUT and the previous continuous speech-based S2ST model \cite{popuri2022enhanced} by finetuning them on text inputs with T2ST task. Both models are pre-trained to perform many-to-many S2ST as described in Table \ref{table:3} and Table \ref{table:3comet}, and transferred to T2ST with the same training configuration (\ie, learning rate, batch size, etc.). We evaluate the T2ST performances of each model for every 10k finetuning steps on the validation translation sets of Es-En, En-Fr, It-Es, and De-Nl. The results are presented in Fig. \ref{fig:4}. We can confirm that the proposed UTUT model transfers to text-based systems more rapidly than the previous continuous speech-based S2ST model. Especially, the performances of UTUT are greatly improved between 10k and 20k finetuning steps. This efficiency is attributed to the UTUT model being trained using a limited set (\ie, vocabulary size) of discretized speech representations, the speech units, which primarily hold the phonetic information. In contrast, the continuous speech-based model requires more time to establish the mapping between text and the pre-learned speech latent space, as this latent space is constructed using an unlimited range of continuous input values.

%------------------------------------ Table 10
%#############################################
\begin{table*}[t]
\renewcommand{\arraystretch}{1.1}
\renewcommand{\tabcolsep}{3.5mm}
\centering
\caption{Intelligibility (CER, \%) evaluation of T2S system on LRS3, mTEDx, and Europarl-ST.}
\label{table:8}
\resizebox{0.7\linewidth}{!}{
\begin{tabular}{ccccccccccc}
\Xhline{3\arrayrulewidth}
\multirow{2}{*}{\textbf{Method}} & \textbf{LRS3} & \multicolumn{3}{c}{\textbf{mTEDx}} & \multicolumn{6}{c}{\textbf{Europarl-ST}}\\ \cmidrule(l{2pt}r{2pt}){2-2} \cmidrule(l{2pt}r{2pt}){3-5} \cmidrule(l{2pt}r{2pt}){6-11}
& \textbf{En} & \textbf{Es} & \textbf{Fr} & \textbf{It}  & \textbf{En} & \textbf{Es} & \textbf{Fr} & \textbf{It} & \textbf{De} & \textbf{Nl} \\ \cmidrule(l{2pt}r{2pt}){1-1}\cmidrule(l{2pt}r{2pt}){2-2}\cmidrule(l{2pt}r{2pt}){3-5} \cmidrule(l{2pt}r{2pt}){6-11}
ReSyn  & 10.3 & 10.9 & 14.0 & 13.2 & 13.6 & 9.8 & 9.3 & 12.4 & 16.9 & 17.7 \\ 
\textbf{UTUT} & 4.3 & 6.6 & 7.2 & 9.8 & 6.2 & 9.5 & 8.1 & 12.8 & 14.4 & 13.9 \\
\Xhline{3\arrayrulewidth}
\end{tabular}}
\end{table*}
%#############################################

%------------------------------------ Table 11
%#############################################
\begin{table}[t]
\renewcommand{\arraystretch}{1.2}
\renewcommand{\tabcolsep}{2.5mm}
\centering
\caption{S2ST BLEU score comparison with denoising pre-training strategy of mBART. The models for both methods are initialized from scratch and trained using the same training configuration.}
\label{table:9}
\resizebox{0.9\linewidth}{!}{
\begin{tabular}{ccccc}
\Xhline{3\arrayrulewidth}
\textbf{Method} & \textbf{Es-En} & \textbf{En-Es} & \textbf{It-Fr} & \textbf{Pl-Fr} \\ \hline
mBART \cite{liu2020mbart,popuri2022enhanced} & 18.6 & 20.2 & 14.2 & 7.4 \\ 
\textbf{UTUT} & \textbf{20.3} & \textbf{22.4} & \textbf{15.4} & \textbf{11.0} \\ \Xhline{3\arrayrulewidth}
\end{tabular}}
\end{table}
%#############################################

\subsection{Comparison with denoising pre-training method}
The denoising pre-training of mBART \cite{liu2020mbart,popuri2022enhanced} is popular for its effectiveness when the pre-trained model is finetuned for language translation tasks. Specifically, it has the advantage that its pre-training does not require a parallel corpus. As the proposed UTUT requires a parallel corpus, we aim to compare the performances of the two strategies, mBART and UTUT, in S2ST, to determine whether better performance can be achieved by employing a parallel corpus during pre-training.
To this end, we pre-train models using different strategies (\ie, mBART and UTUT) on the same pre-training dataset described in Sec. \ref{sec:4.1}. Both models have the same architecture and are initialized from scratch. After pre-training with 500k steps, we finetune both models for 30k steps on each target language pair of En-Es, Es-En, It-Fr, and Pl-Fr, to assess how different pre-training methods affect the final S2ST performances. The best results of each method are shown in Table \ref{table:9}.
The results confirm that better performances can be achieved by pre-training the model with a many-to-many language translation task, compared to the denoising pre-training strategy. Therefore, if access to a parallel corpus is available, it would be beneficial to employ the UTUT training strategy instead of the denoising pre-training strategy. Especially for the language pair with fewer training data, Pl-Fr, the proposed UTUT outperforms mBART by a larger margin. This indicates that building associations between different languages during pre-training is beneficial. It is important to note that through the many-to-many language translation, the proposed UTUT model can even extend its knowledge to translate between unseen language pairs.

\subsection{Effectiveness of input masking strategy}
We perform an ablation study to confirm whether input masking strategies are beneficial for UTUT pre-training. To this end, we build 4 variants of models trained with the masking percentage $p_m$ of 0, 10, 30, and 50, for 200k steps. The results are shown in Table \ref{table:10}. We observe that input masking strategies do not result in improved performance in unit-to-unit translation, unlike their effectiveness in reconstruction tasks \cite{devlin2018bert,lewis2019bart}. Using 0\% masking consistently achieves better performance across various language pair conditions, therefore we empirically use the model for the other experiments. 

%------------------------------------ Table 12
%#############################################
\begin{table}[t]
\renewcommand{\arraystretch}{1.2}
\renewcommand{\tabcolsep}{4mm}
\centering
\caption{S2ST ablation results (BLEU) on different percent $p_m$ for input masking strategies at 200k steps.}
\label{table:10}
\resizebox{0.8\linewidth}{!}{
\begin{tabular}[b]{ccccc}
\Xhline{3\arrayrulewidth}
$p_m$ & \textbf{Es-En} & \textbf{En-Es} & \textbf{It-Fr} & \textbf{Pl-Fr}  \\ \hline
-    & \textbf{20.08} & \textbf{21.83} & \textbf{14.01} & 9.93 \\
10\% & 19.32 & 21.53 & 13.92 & \textbf{10.05} \\ 
30\% & 18.72 & 20.30 & 13.60 & 9.59 \\ 
50\% & 18.39 & 20.88 & 13.52 & 9.30 \\  \Xhline{3\arrayrulewidth}
\end{tabular}}
\end{table}
%#############################################

\subsection{Intelligibility of the T2S system}
In order to evaluate the intelligibility of the T2S system, we measure Character Error Rate (CER) by using a pre-trained ASR model. As the proposed system generates speech from speech units, we set the baseline performance as the speech resynthesis quality of the target speech units, following \cite{lakhotia2021generative}. Therefore, we evaluate the CER of generated speech of UTUT by comparing it with the resynthesized performance of ground truth speech units. The performances of baseline (`ReSyn' in the table) and UTUT are shown in Table \ref{table:8}. The results show that the generated speech from UTUT using text inputs is quite intelligible by achieving better CER than the resynthesized speech overall. It means that by generating the speech from text, the model can generate the target speech by focusing on the linguistic content only while eliminating undesired sounds (\eg, noise) from the speech, which may be residue in the ground truth speech (\ie, ground truth speech units).

\section{Conclusion}
We have proposed a textless multilingual many-to-many speech-to-speech training method, Unit-to-Unit Translation (UTUT). The proposed UTUT is designed in a many-to-many spoken language translation setting so that we can effectively train a speech-to-speech machine translation system with much less training data compared to a unidirectional translation system. We showed that even unseen language pairs can be translated by the proposed UTUT model. Moreover, by associating the speech units with text modality, we can easily transfer the trained UTUT model to handle text inputs. Through extensive experiments on S2ST, T2S, and T2ST using diverse languages, we validated the effectiveness of the proposed UTUT in multilingual speech and text processing.

\bibliographystyle{IEEEtran}
\bibliography{main}

\begin{thebibliography}{100}
\providecommand{\url}[1]{#1}
\csname url@samestyle\endcsname
\providecommand{\newblock}{\relax}
\providecommand{\bibinfo}[2]{#2}
\providecommand{\BIBentrySTDinterwordspacing}{\spaceskip=0pt\relax}
\providecommand{\BIBentryALTinterwordstretchfactor}{4}
\providecommand{\BIBentryALTinterwordspacing}{\spaceskip=\fontdimen2\font plus
\BIBentryALTinterwordstretchfactor\fontdimen3\font minus \fontdimen4\font\relax}
\providecommand{\BIBforeignlanguage}[2]{{%
\expandafter\ifx\csname l@#1\endcsname\relax
\typeout{** WARNING: IEEEtran.bst: No hyphenation pattern has been}%
\typeout{** loaded for the language `#1'. Using the pattern for}%
\typeout{** the default language instead.}%
\else
\language=\csname l@#1\endcsname
\fi
#2}}
\providecommand{\BIBdecl}{\relax}
\BIBdecl

\bibitem{hannun2014deepspeech}
A.~Hannun, C.~Case, J.~Casper, B.~Catanzaro, G.~Diamos, E.~Elsen, R.~Prenger, S.~Satheesh, S.~Sengupta, A.~Coates \emph{et~al.}, ``Deep speech: Scaling up end-to-end speech recognition,'' \emph{arXiv preprint arXiv:1412.5567}, 2014.

\bibitem{amodei2016deep}
D.~Amodei, S.~Ananthanarayanan, R.~Anubhai, J.~Bai, E.~Battenberg, C.~Case, J.~Casper, B.~Catanzaro, Q.~Cheng, G.~Chen \emph{et~al.}, ``Deep speech 2: End-to-end speech recognition in english and mandarin,'' in \emph{International conference on machine learning}.\hskip 1em plus 0.5em minus 0.4em\relax PMLR, 2016, pp. 173--182.

\bibitem{kim2017joint}
S.~Kim, T.~Hori, and S.~Watanabe, ``Joint ctc-attention based end-to-end speech recognition using multi-task learning,'' in \emph{2017 IEEE international conference on acoustics, speech and signal processing (ICASSP)}.\hskip 1em plus 0.5em minus 0.4em\relax IEEE, 2017, pp. 4835--4839.

\bibitem{watanabe2017hybrid}
S.~Watanabe, T.~Hori, S.~Kim, J.~R. Hershey, and T.~Hayashi, ``Hybrid ctc/attention architecture for end-to-end speech recognition,'' \emph{IEEE Journal of Selected Topics in Signal Processing}, vol.~11, no.~8, pp. 1240--1253, 2017.

\bibitem{watanabe2018espnet}
S.~Watanabe, T.~Hori, S.~Karita, T.~Hayashi, J.~Nishitoba, Y.~Unno, N.~E.~Y. Soplin, J.~Heymann, M.~Wiesner, N.~Chen \emph{et~al.}, ``Espnet: End-to-end speech processing toolkit,'' \emph{arXiv preprint arXiv:1804.00015}, 2018.

\bibitem{kim2021cromm}
M.~Kim, J.~Hong, S.~J. Park, and Y.~M. Ro, ``Cromm-vsr: Cross-modal memory augmented visual speech recognition,'' \emph{IEEE Transactions on Multimedia}, vol.~24, pp. 4342--4355, 2021.

\bibitem{shi2022avhubert}
B.~Shi, W.-N. Hsu, K.~Lakhotia, and A.~Mohamed, ``Learning audio-visual speech representation by masked multimodal cluster prediction,'' \emph{arXiv preprint arXiv:2201.02184}, 2022.

\bibitem{kim2022distinguishing}
M.~Kim, J.~H. Yeo, and Y.~M. Ro, ``Distinguishing homophenes using multi-head visual-audio memory for lip reading,'' in \emph{Proceedings of the AAAI Conference on Artificial Intelligence}, vol.~36, no.~1, 2022, pp. 1174--1182.

\bibitem{hong2023watch}
J.~Hong, M.~Kim, J.~Choi, and Y.~M. Ro, ``Watch or listen: Robust audio-visual speech recognition with visual corruption modeling and reliability scoring,'' \emph{arXiv preprint arXiv:2303.08536}, 2023.

\bibitem{bahdanau2014neural}
D.~Bahdanau, K.~Cho, and Y.~Bengio, ``Neural machine translation by jointly learning to align and translate,'' \emph{arXiv preprint arXiv:1409.0473}, 2014.

\bibitem{wu2016google}
Y.~Wu, M.~Schuster, Z.~Chen, Q.~V. Le, M.~Norouzi, W.~Macherey, M.~Krikun, Y.~Cao, Q.~Gao, K.~Macherey \emph{et~al.}, ``Google's neural machine translation system: Bridging the gap between human and machine translation,'' \emph{arXiv preprint arXiv:1609.08144}, 2016.

\bibitem{chen2018best}
M.~X. Chen, O.~Firat, A.~Bapna, M.~Johnson, W.~Macherey, G.~Foster, L.~Jones, N.~Parmar, M.~Schuster, Z.~Chen \emph{et~al.}, ``The best of both worlds: Combining recent advances in neural machine translation,'' \emph{arXiv preprint arXiv:1804.09849}, 2018.

\bibitem{brown2020gpt-3}
T.~Brown, B.~Mann, N.~Ryder, M.~Subbiah, J.~D. Kaplan, P.~Dhariwal, A.~Neelakantan, P.~Shyam, G.~Sastry, A.~Askell \emph{et~al.}, ``Language models are few-shot learners,'' \emph{Advances in neural information processing systems}, vol.~33, pp. 1877--1901, 2020.

\bibitem{liu2020mbart}
Y.~Liu, J.~Gu, N.~Goyal, X.~Li, S.~Edunov, M.~Ghazvininejad, M.~Lewis, and L.~Zettlemoyer, ``Multilingual denoising pre-training for neural machine translation,'' \emph{Transactions of the Association for Computational Linguistics}, vol.~8, pp. 726--742, 2020.

\bibitem{wang2017tacotron}
Y.~Wang, R.~Skerry-Ryan, D.~Stanton, Y.~Wu, R.~J. Weiss, N.~Jaitly, Z.~Yang, Y.~Xiao, Z.~Chen, S.~Bengio \emph{et~al.}, ``Tacotron: A fully end-to-end text-to-speech synthesis model,'' \emph{arXiv preprint arXiv:1703.10135}, vol. 164, 2017.

\bibitem{jia2018transfer}
Y.~Jia, Y.~Zhang, R.~Weiss, Q.~Wang, J.~Shen, F.~Ren, P.~Nguyen, R.~Pang, I.~Lopez~Moreno, Y.~Wu \emph{et~al.}, ``Transfer learning from speaker verification to multispeaker text-to-speech synthesis,'' \emph{Advances in neural information processing systems}, vol.~31, 2018.

\bibitem{valle2020flowtron}
R.~Valle, K.~Shih, R.~Prenger, and B.~Catanzaro, ``Flowtron: an autoregressive flow-based generative network for text-to-speech synthesis,'' \emph{arXiv preprint arXiv:2005.05957}, 2020.

\bibitem{chen2021adaspeech}
M.~Chen, X.~Tan, B.~Li, Y.~Liu, T.~Qin, S.~Zhao, and T.-Y. Liu, ``Adaspeech: Adaptive text to speech for custom voice,'' \emph{arXiv preprint arXiv:2103.00993}, 2021.

\bibitem{casanova2022yourtts}
E.~Casanova, J.~Weber, C.~D. Shulby, A.~C. Junior, E.~G{\"o}lge, and M.~A. Ponti, ``Yourtts: Towards zero-shot multi-speaker tts and zero-shot voice conversion for everyone,'' in \emph{International Conference on Machine Learning}, 2022, pp. 2709--2720.

\bibitem{wang2023neural}
C.~Wang, S.~Chen, Y.~Wu, Z.~Zhang, L.~Zhou, S.~Liu, Z.~Chen, Y.~Liu, H.~Wang, J.~Li \emph{et~al.}, ``Neural codec language models are zero-shot text to speech synthesizers,'' \emph{arXiv preprint arXiv:2301.02111}, 2023.

\bibitem{thoppilan2022lamda}
R.~Thoppilan, D.~De~Freitas, J.~Hall, N.~Shazeer, A.~Kulshreshtha, H.-T. Cheng, A.~Jin, T.~Bos, L.~Baker, Y.~Du \emph{et~al.}, ``Lamda: Language models for dialog applications,'' \emph{arXiv preprint arXiv:2201.08239}, 2022.

\bibitem{touvron2023llama}
H.~Touvron, T.~Lavril, G.~Izacard, X.~Martinet, M.-A. Lachaux, T.~Lacroix, B.~Rozi{\`e}re, N.~Goyal, E.~Hambro, F.~Azhar \emph{et~al.}, ``Llama: Open and efficient foundation language models,'' \emph{arXiv preprint arXiv:2302.13971}, 2023.

\bibitem{chatgpt}
OpenAI, ``Openai: Introducing chatgpt,'' URL \url{https://openai.com/blog/chatgpt}, 2022.

\bibitem{zhang2019learning}
Y.~Zhang, R.~J. Weiss, H.~Zen, Y.~Wu, Z.~Chen, R.~Skerry-Ryan, Y.~Jia, A.~Rosenberg, and B.~Ramabhadran, ``Learning to speak fluently in a foreign language: Multilingual speech synthesis and cross-language voice cloning,'' \emph{arXiv preprint arXiv:1907.04448}, 2019.

\bibitem{nekvinda2020one}
T.~Nekvinda and O.~Du{\v{s}}ek, ``One model, many languages: Meta-learning for multilingual text-to-speech,'' \emph{arXiv preprint arXiv:2008.00768}, 2020.

\bibitem{zhang2023valle-x}
Z.~Zhang, L.~Zhou, C.~Wang, S.~Chen, Y.~Wu, S.~Liu, Z.~Chen, Y.~Liu, H.~Wang, J.~Li \emph{et~al.}, ``Speak foreign languages with your own voice: Cross-lingual neural codec language modeling,'' \emph{arXiv preprint arXiv:2303.03926}, 2023.

\bibitem{saeki2023learning}
T.~Saeki, S.~Maiti, X.~Li, S.~Watanabe, S.~Takamichi, and H.~Saruwatari, ``Learning to speak from text: Zero-shot multilingual text-to-speech with unsupervised text pretraining,'' \emph{arXiv preprint arXiv:2301.12596}, 2023.

\bibitem{sennrich2015bytepair}
R.~Sennrich, B.~Haddow, and A.~Birch, ``Neural machine translation of rare words with subword units,'' \emph{arXiv preprint arXiv:1508.07909}, 2015.

\bibitem{kudo2018subword}
T.~Kudo, ``Subword regularization: Improving neural network translation models with multiple subword candidates,'' \emph{arXiv preprint arXiv:1804.10959}, 2018.

\bibitem{kudo2018sentencepiece}
T.~Kudo and J.~Richardson, ``Sentencepiece: A simple and language independent subword tokenizer and detokenizer for neural text processing,'' \emph{arXiv preprint arXiv:1808.06226}, 2018.

\bibitem{Bernard2021}
\BIBentryALTinterwordspacing
M.~Bernard and H.~Titeux, ``Phonemizer: Text to phones transcription for multiple languages in python,'' \emph{Journal of Open Source Software}, vol.~6, no.~68, p. 3958, 2021. [Online]. Available: \url{https://doi.org/10.21105/joss.03958}
\BIBentrySTDinterwordspacing

\bibitem{lakhotia2021generative}
K.~Lakhotia, E.~Kharitonov, W.-N. Hsu, Y.~Adi, A.~Polyak, B.~Bolte, T.-A. Nguyen, J.~Copet, A.~Baevski, A.~Mohamed \emph{et~al.}, ``On generative spoken language modeling from raw audio,'' \emph{Transactions of the Association for Computational Linguistics}, vol.~9, pp. 1336--1354, 2021.

\bibitem{polyak2021speech}
A.~Polyak, Y.~Adi, J.~Copet, E.~Kharitonov, K.~Lakhotia, W.-N. Hsu, A.~Mohamed, and E.~Dupoux, ``Speech resynthesis from discrete disentangled self-supervised representations,'' \emph{arXiv preprint arXiv:2104.00355}, 2021.

\bibitem{choi2023intelligible}
J.~Choi, M.~Kim, and Y.~M. Ro, ``Intelligible lip-to-speech synthesis with speech units,'' \emph{arXiv preprint arXiv:2305.19603}, 2023.

\bibitem{hsu2021hubert}
W.-N. Hsu, B.~Bolte, Y.-H.~H. Tsai, K.~Lakhotia, R.~Salakhutdinov, and A.~Mohamed, ``Hubert: Self-supervised speech representation learning by masked prediction of hidden units,'' \emph{IEEE/ACM Transactions on Audio, Speech, and Language Processing}, vol.~29, pp. 3451--3460, 2021.

\bibitem{chang2023exploration}
X.~Chang, B.~Yan, Y.~Fujita, T.~Maekaku, and S.~Watanabe, ``Exploration of efficient end-to-end asr using discretized input from self-supervised learning,'' \emph{arXiv preprint arXiv:2305.18108}, 2023.

\bibitem{baevski2020wav2vec}
A.~Baevski, Y.~Zhou, A.~Mohamed, and M.~Auli, ``wav2vec 2.0: A framework for self-supervised learning of speech representations,'' \emph{Advances in neural information processing systems}, vol.~33, pp. 12\,449--12\,460, 2020.

\bibitem{babu2021xls}
A.~Babu, C.~Wang, A.~Tjandra, K.~Lakhotia, Q.~Xu, N.~Goyal, K.~Singh, P.~von Platen, Y.~Saraf, J.~Pino \emph{et~al.}, ``Xls-r: Self-supervised cross-lingual speech representation learning at scale,'' \emph{arXiv preprint arXiv:2111.09296}, 2021.

\bibitem{borsos2023audiolm}
Z.~Borsos, R.~Marinier, D.~Vincent, E.~Kharitonov, O.~Pietquin, M.~Sharifi, D.~Roblek, O.~Teboul, D.~Grangier, M.~Tagliasacchi \emph{et~al.}, ``Audiolm: a language modeling approach to audio generation,'' \emph{IEEE/ACM Transactions on Audio, Speech, and Language Processing}, 2023.

\bibitem{schultz2001multilingualASR}
T.~Schultz and A.~Waibel, ``Language-independent and language-adaptive acoustic modeling for speech recognition,'' \emph{Speech Communication}, vol.~35, no. 1-2, pp. 31--51, 2001.

\bibitem{vu2014multilingual}
N.~T. Vu, D.~Imseng, D.~Povey, P.~Motlicek, T.~Schultz, and H.~Bourlard, ``Multilingual deep neural network based acoustic modeling for rapid language adaptation,'' in \emph{2014 IEEE international Conference on acoustics, speech and signal processing (ICASSP)}.\hskip 1em plus 0.5em minus 0.4em\relax IEEE, 2014, pp. 7639--7643.

\bibitem{luo2020multilingualVSR}
M.~Luo, S.~Yang, X.~Chen, Z.~Liu, and S.~Shan, ``Synchronous bidirectional learning for multilingual lip reading,'' \emph{arXiv preprint arXiv:2005.03846}, 2020.

\bibitem{lee2021textless}
A.~Lee, H.~Gong, P.-A. Duquenne, H.~Schwenk, P.-J. Chen, C.~Wang, S.~Popuri, Y.~Adi, J.~Pino, J.~Gu \emph{et~al.}, ``Textless speech-to-speech translation on real data,'' in \emph{Proceedings of the 2022 Conference of the North American Chapter of the Association for Computational Linguistics: Human Language Technologies}, 2022, pp. 860--872.

\bibitem{chen2022wavlm}
S.~Chen, C.~Wang, Z.~Chen, Y.~Wu, S.~Liu, Z.~Chen, J.~Li, N.~Kanda, T.~Yoshioka, X.~Xiao \emph{et~al.}, ``Wavlm: Large-scale self-supervised pre-training for full stack speech processing,'' \emph{IEEE Journal of Selected Topics in Signal Processing}, vol.~16, no.~6, pp. 1505--1518, 2022.

\bibitem{ao2021speecht5}
J.~Ao, R.~Wang, L.~Zhou, C.~Wang, S.~Ren, Y.~Wu, S.~Liu, T.~Ko, Q.~Li, Y.~Zhang \emph{et~al.}, ``Speecht5: Unified-modal encoder-decoder pre-training for spoken language processing,'' \emph{arXiv preprint arXiv:2110.07205}, 2021.

\bibitem{chung2017lip}
J.~S. Chung, A.~Senior, O.~Vinyals, and A.~Zisserman, ``Lip reading sentences in the wild,'' in \emph{2017 IEEE conference on computer vision and pattern recognition (CVPR)}.\hskip 1em plus 0.5em minus 0.4em\relax IEEE, 2017, pp. 3444--3453.

\bibitem{hong2021speech}
J.~Hong, M.~Kim, S.~J. Park, and Y.~M. Ro, ``Speech reconstruction with reminiscent sound via visual voice memory,'' \emph{IEEE/ACM Transactions on Audio, Speech, and Language Processing}, vol.~29, pp. 3654--3667, 2021.

\bibitem{kim2023prompt}
M.~Kim, H.-I. Kim, and Y.~M. Ro, ``Prompt tuning of deep neural networks for speaker-adaptive visual speech recognition,'' \emph{arXiv preprint arXiv:2302.08102}, 2023.

\bibitem{yeo2023multi}
J.~H. Yeo, M.~Kim, and Y.~M. Ro, ``Multi-temporal lip-audio memory for visual speech recognition,'' in \emph{ICASSP 2023-2023 IEEE International Conference on Acoustics, Speech and Signal Processing (ICASSP)}.\hskip 1em plus 0.5em minus 0.4em\relax IEEE, 2023, pp. 1--5.

\bibitem{chang2023exploring}
\emph{Exploring Speech Recognition, Translation, and Understanding with Discrete Speech Units: A Comparative Study}, 2024.

\bibitem{nguyen2023generative}
T.~A. Nguyen, E.~Kharitonov, J.~Copet, Y.~Adi, W.-N. Hsu, A.~Elkahky, P.~Tomasello, R.~Algayres, B.~Sagot, A.~Mohamed \emph{et~al.}, ``Generative spoken dialogue language modeling,'' \emph{Transactions of the Association for Computational Linguistics}, vol.~11, pp. 250--266, 2023.

\bibitem{kreuk2021textlessemotion}
F.~Kreuk, A.~Polyak, J.~Copet, E.~Kharitonov, T.-A. Nguyen, M.~Rivi{\`e}re, W.-N. Hsu, A.~Mohamed, E.~Dupoux, and Y.~Adi, ``Textless speech emotion conversion using decomposed and discrete representations,'' \emph{arXiv preprint arXiv:2111.07402}, 2021.

\bibitem{do2016preserving}
Q.~T. Do, T.~Toda, G.~Neubig, S.~Sakti, and S.~Nakamura, ``Preserving word-level emphasis in speech-to-speech translation,'' \emph{IEEE/ACM Transactions on Audio, Speech, and Language Processing}, vol.~25, no.~3, pp. 544--556, 2016.

\bibitem{li2014overview}
J.~Li, L.~Deng, Y.~Gong, and R.~Haeb-Umbach, ``An overview of noise-robust automatic speech recognition,'' \emph{IEEE/ACM Transactions on Audio, Speech, and Language Processing}, vol.~22, no.~4, pp. 745--777, 2014.

\bibitem{abdel2014convolutional}
O.~Abdel-Hamid, A.-r. Mohamed, H.~Jiang, L.~Deng, G.~Penn, and D.~Yu, ``Convolutional neural networks for speech recognition,'' \emph{IEEE/ACM Transactions on audio, speech, and language processing}, vol.~22, no.~10, pp. 1533--1545, 2014.

\bibitem{zhang2020improving}
W.~Zhang, X.~Chang, Y.~Qian, and S.~Watanabe, ``Improving end-to-end single-channel multi-talker speech recognition,'' \emph{IEEE/ACM Transactions on Audio, Speech, and Language Processing}, vol.~28, pp. 1385--1394, 2020.

\bibitem{su2018hierarchy}
J.~Su, J.~Zeng, D.~Xiong, Y.~Liu, M.~Wang, and J.~Xie, ``A hierarchy-to-sequence attentional neural machine translation model,'' \emph{IEEE/ACM Transactions on Audio, Speech, and Language Processing}, vol.~26, no.~3, pp. 623--632, 2018.

\bibitem{yang2022gtrans}
J.~Yang, Y.~Yin, L.~Yang, S.~Ma, H.~Huang, D.~Zhang, F.~Wei, and Z.~Li, ``Gtrans: Grouping and fusing transformer layers for neural machine translation,'' \emph{IEEE/ACM Transactions on Audio, Speech, and Language Processing}, vol.~31, pp. 1489--1498, 2022.

\bibitem{huang2022meta}
S.-F. Huang, C.-J. Lin, D.-R. Liu, Y.-C. Chen, and H.-y. Lee, ``Meta-tts: Meta-learning for few-shot speaker adaptive text-to-speech,'' \emph{IEEE/ACM Transactions on Audio, Speech, and Language Processing}, vol.~30, pp. 1558--1571, 2022.

\bibitem{du2023speaker}
C.~Du, Y.~Guo, X.~Chen, and K.~Yu, ``Speaker adaptive text-to-speech with timbre-normalized vector-quantized feature,'' \emph{IEEE/ACM Transactions on Audio, Speech, and Language Processing}, 2023.

\bibitem{jia2019translatotron}
Y.~Jia, R.~J. Weiss, F.~Biadsy, W.~Macherey, M.~Johnson, Z.~Chen, and Y.~Wu, ``Direct speech-to-speech translation with a sequence-to-sequence model,'' \emph{arXiv preprint arXiv:1904.06037}, 2019.

\bibitem{kano2020end}
T.~Kano, S.~Sakti, and S.~Nakamura, ``End-to-end speech translation with transcoding by multi-task learning for distant language pairs,'' \emph{IEEE/ACM Transactions on Audio, Speech, and Language Processing}, vol.~28, pp. 1342--1355, 2020.

\bibitem{jia2021translatotron2}
Y.~Jia, M.~T. Ramanovich, T.~Remez, and R.~Pomerantz, ``Translatotron 2: High-quality direct speech-to-speech translation with voice preservation,'' in \emph{International Conference on Machine Learning}.\hskip 1em plus 0.5em minus 0.4em\relax PMLR, 2022, pp. 10\,120--10\,134.

\bibitem{kano2021transformer}
T.~Kano, S.~Sakti, and S.~Nakamura, ``Transformer-based direct speech-to-speech translation with transcoder,'' in \emph{2021 IEEE Spoken Language Technology Workshop (SLT)}.\hskip 1em plus 0.5em minus 0.4em\relax IEEE, 2021, pp. 958--965.

\bibitem{vaswani2017attention}
A.~Vaswani, N.~Shazeer, N.~Parmar, J.~Uszkoreit, L.~Jones, A.~N. Gomez, {\L}.~Kaiser, and I.~Polosukhin, ``Attention is all you need,'' \emph{Advances in neural information processing systems}, vol.~30, 2017.

\bibitem{huang2022transpeech}
R.~Huang, Z.~Zhao, J.~Liu, H.~Liu, Y.~Ren, L.~Zhang, and J.~He, ``Transpeech: Speech-to-speech translation with bilateral perturbation,'' \emph{arXiv preprint arXiv:2205.12523}, 2022.

\bibitem{inaguma2022unity}
H.~Inaguma, S.~Popuri, I.~Kulikov, P.-J. Chen, C.~Wang, Y.-A. Chung, Y.~Tang, A.~Lee, S.~Watanabe, and J.~Pino, ``Unity: Two-pass direct speech-to-speech translation with discrete units,'' \emph{arXiv preprint arXiv:2212.08055}, 2022.

\bibitem{lee2021directs2s}
A.~Lee, P.-J. Chen, C.~Wang, J.~Gu, S.~Popuri, X.~Ma, A.~Polyak, Y.~Adi, Q.~He, Y.~Tang \emph{et~al.}, ``Direct speech-to-speech translation with discrete units,'' in \emph{Proceedings of the 60th Annual Meeting of the Association for Computational Linguistics (Volume 1: Long Papers)}, 2022.

\bibitem{jia2022leveraging}
Y.~Jia, Y.~Ding, A.~Bapna, C.~Cherry, Y.~Zhang, A.~Conneau, and N.~Morioka, ``Leveraging unsupervised and weakly-supervised data to improve direct speech-to-speech translation,'' \emph{arXiv preprint arXiv:2203.13339}, 2022.

\bibitem{chung2021w2v}
Y.-A. Chung, Y.~Zhang, W.~Han, C.-C. Chiu, J.~Qin, R.~Pang, and Y.~Wu, ``W2v-bert: Combining contrastive learning and masked language modeling for self-supervised speech pre-training,'' in \emph{2021 IEEE Automatic Speech Recognition and Understanding Workshop (ASRU)}.\hskip 1em plus 0.5em minus 0.4em\relax IEEE, 2021, pp. 244--250.

\bibitem{popuri2022enhanced}
S.~Popuri, P.-J. Chen, C.~Wang, J.~Pino, Y.~Adi, J.~Gu, W.-N. Hsu, and A.~Lee, ``Enhanced direct speech-to-speech translation using self-supervised pre-training and data augmentation,'' \emph{arXiv preprint arXiv:2204.02967}, 2022.

\bibitem{diwan2024textless}
A.~Diwan, A.~Srinivasan, D.~Harwath, and E.~Choi, ``Textless low-resource speech-to-speech translation with unit language models,'' 2024.

\bibitem{sennrich2016improving}
R.~Sennrich, B.~Haddow, and A.~Birch, ``Improving neural machine translation models with monolingual data,'' in \emph{Proceedings of the 54th Annual Meeting of the Association for Computational Linguistics (Volume 1: Long Papers)}.\hskip 1em plus 0.5em minus 0.4em\relax Association for Computational Linguistics, 2016.

\bibitem{barrault2023seamlessm4t}
L.~Barrault, Y.-A. Chung, M.~C. Meglioli, D.~Dale, N.~Dong, P.-A. Duquenne, H.~Elsahar, H.~Gong, K.~Heffernan, J.~Hoffman \emph{et~al.}, ``Seamlessm4t-massively multilingual \& multimodal machine translation,'' \emph{arXiv preprint arXiv:2308.11596}, 2023.

\bibitem{ren2019fastspeech}
Y.~Ren, Y.~Ruan, X.~Tan, T.~Qin, S.~Zhao, Z.~Zhao, and T.-Y. Liu, ``Fastspeech: Fast, robust and controllable text to speech,'' \emph{Advances in neural information processing systems}, vol.~32, 2019.

\bibitem{arik2017deep}
S.~{\"O}. Ar{\i}k, M.~Chrzanowski, A.~Coates, G.~Diamos, A.~Gibiansky, Y.~Kang, X.~Li, J.~Miller, A.~Ng, J.~Raiman \emph{et~al.}, ``Deep voice: Real-time neural text-to-speech,'' in \emph{International conference on machine learning}, 2017, pp. 195--204.

\bibitem{cooper2020zero}
E.~Cooper, C.-I. Lai, Y.~Yasuda, F.~Fang, X.~Wang, N.~Chen, and J.~Yamagishi, ``Zero-shot multi-speaker text-to-speech with state-of-the-art neural speaker embeddings,'' in \emph{ICASSP 2020-2020 IEEE International Conference on Acoustics, Speech and Signal Processing (ICASSP)}.\hskip 1em plus 0.5em minus 0.4em\relax IEEE, 2020, pp. 6184--6188.

\bibitem{azizah2020hierarchical}
K.~Azizah, M.~Adriani, and W.~Jatmiko, ``Hierarchical transfer learning for multilingual, multi-speaker, and style transfer dnn-based tts on low-resource languages,'' \emph{IEEE Access}, vol.~8, pp. 179\,798--179\,812, 2020.

\bibitem{xu2020lrspeech}
J.~Xu, X.~Tan, Y.~Ren, T.~Qin, J.~Li, S.~Zhao, and T.-Y. Liu, ``Lrspeech: Extremely low-resource speech synthesis and recognition,'' in \emph{Proceedings of the 26th ACM SIGKDD International Conference on Knowledge Discovery \& Data Mining}, 2020, pp. 2802--2812.

\bibitem{he2021multilingual}
M.~He, J.~Yang, L.~He, and F.~K. Soong, ``Multilingual byte2speech models for scalable low-resource speech synthesis,'' \emph{arXiv preprint arXiv:2103.03541}, 2021.

\bibitem{lux2022low}
F.~Lux, J.~Koch, and N.~T. Vu, ``Low-resource multilingual and zero-shot multispeaker tts,'' \emph{arXiv preprint arXiv:2210.12223}, 2022.

\bibitem{defossez2022high}
A.~D{\'e}fossez, J.~Copet, G.~Synnaeve, and Y.~Adi, ``High fidelity neural audio compression,'' \emph{arXiv preprint arXiv:2210.13438}, 2022.

\bibitem{sicherman2023analysing}
A.~Sicherman and Y.~Adi, ``Analysing discrete self supervised speech representation for spoken language modeling,'' in \emph{ICASSP 2023-2023 IEEE International Conference on Acoustics, Speech and Signal Processing (ICASSP)}.\hskip 1em plus 0.5em minus 0.4em\relax IEEE, 2023, pp. 1--5.

\bibitem{zhang2022speechut}
Z.~Zhang, L.~Zhou, J.~Ao, S.~Liu, L.~Dai, J.~Li, and F.~Wei, ``Speechut: Bridging speech and text with hidden-unit for encoder-decoder based speech-text pre-training,'' \emph{arXiv preprint arXiv:2210.03730}, 2022.

\bibitem{li2022textlesss2s}
X.~Li, Y.~Jia, and C.-C. Chiu, ``Textless direct speech-to-speech translation with discrete speech representation,'' in \emph{ICASSP 2023-2023 IEEE International Conference on Acoustics, Speech and Signal Processing (ICASSP)}.\hskip 1em plus 0.5em minus 0.4em\relax IEEE, 2023, pp. 1--5.

\bibitem{wang2021voxpopuli}
C.~Wang, M.~Riviere, A.~Lee, A.~Wu, C.~Talnikar, D.~Haziza, M.~Williamson, J.~Pino, and E.~Dupoux, ``Voxpopuli: A large-scale multilingual speech corpus for representation learning, semi-supervised learning and interpretation,'' \emph{arXiv preprint arXiv:2101.00390}, 2021.

\bibitem{lewis2019bart}
M.~Lewis, Y.~Liu, N.~Goyal, M.~Ghazvininejad, A.~Mohamed, O.~Levy, V.~Stoyanov, and L.~Zettlemoyer, ``Bart: Denoising sequence-to-sequence pre-training for natural language generation, translation, and comprehension,'' \emph{arXiv preprint arXiv:1910.13461}, 2019.

\bibitem{devlin2018bert}
J.~Devlin, M.-W. Chang, K.~Lee, and K.~Toutanova, ``Bert: Pre-training of deep bidirectional transformers for language understanding,'' \emph{arXiv preprint arXiv:1810.04805}, 2018.

\bibitem{ott2019fairseq}
M.~Ott, S.~Edunov, A.~Baevski, A.~Fan, S.~Gross, N.~Ng, D.~Grangier, and M.~Auli, ``fairseq: A fast, extensible toolkit for sequence modeling,'' in \emph{Proceedings of NAACL-HLT 2019: Demonstrations}, 2019.

\bibitem{kong2020hifi}
J.~Kong, J.~Kim, and J.~Bae, ``Hifi-gan: Generative adversarial networks for efficient and high fidelity speech synthesis,'' \emph{Advances in Neural Information Processing Systems}, vol.~33, pp. 17\,022--17\,033, 2020.

\bibitem{salesky2021mtedx}
E.~Salesky, M.~Wiesner, J.~Bremerman, R.~Cattoni, M.~Negri, M.~Turchi, D.~W. Oard, and M.~Post, ``The multilingual tedx corpus for speech recognition and translation,'' \emph{arXiv preprint arXiv:2102.01757}, 2021.

\bibitem{anwar2023muavic}
M.~Anwar, B.~Shi, V.~Goswami, W.-N. Hsu, J.~Pino, and C.~Wang, ``Muavic: A multilingual audio-visual corpus for robust speech recognition and robust speech-to-text translation,'' \emph{arXiv preprint arXiv:2303.00628}, 2023.

\bibitem{kim2021conditional}
J.~Kim, J.~Kong, and J.~Son, ``Conditional variational autoencoder with adversarial learning for end-to-end text-to-speech,'' in \emph{International Conference on Machine Learning}.\hskip 1em plus 0.5em minus 0.4em\relax PMLR, 2021, pp. 5530--5540.

\bibitem{ljspeech17}
K.~Ito and L.~Johnson, ``The lj speech dataset,'' \url{https://keithito.com/LJ-Speech-Dataset/}, 2017.

\bibitem{post-2018-call}
\BIBentryALTinterwordspacing
M.~Post, ``A call for clarity in reporting {BLEU} scores,'' in \emph{Proceedings of the Third Conference on Machine Translation: Research Papers}.\hskip 1em plus 0.5em minus 0.4em\relax Belgium, Brussels: Association for Computational Linguistics, Oct. 2018, pp. 186--191. [Online]. Available: \url{https://www.aclweb.org/anthology/W18-6319}
\BIBentrySTDinterwordspacing

\bibitem{freitag2022comet}
M.~Freitag, R.~Rei, N.~Mathur, C.-k. Lo, C.~Stewart, E.~Avramidis, T.~Kocmi, G.~Foster, A.~Lavie, and A.~F. Martins, ``Results of wmt22 metrics shared task: Stop using bleu--neural metrics are better and more robust,'' in \emph{Proceedings of the Seventh Conference on Machine Translation (WMT)}, 2022, pp. 46--68.

\bibitem{duquenne2022speechmatrix}
P.-A. Duquenne, H.~Gong, N.~Dong, J.~Du, A.~Lee, V.~Goswani, C.~Wang, J.~Pino, B.~Sagot, and H.~Schwenk, ``Speechmatrix: A large-scale mined corpus of multilingual speech-to-speech translations,'' \emph{arXiv preprint arXiv:2211.04508}, 2022.

\bibitem{park2019css10}
K.~Park and T.~Mulc, ``Css10: A collection of single speaker speech datasets for 10 languages,'' \emph{arXiv preprint arXiv:1903.11269}, 2019.

\bibitem{iranzo2020europarl}
J.~Iranzo-S{\'a}nchez, J.~A. Silvestre-Cerda, J.~Jorge, N.~Rosell{\'o}, A.~Gim{\'e}nez, A.~Sanchis, J.~Civera, and A.~Juan, ``Europarl-st: A multilingual corpus for speech translation of parliamentary debates,'' in \emph{ICASSP 2020-2020 IEEE International Conference on Acoustics, Speech and Signal Processing (ICASSP)}.\hskip 1em plus 0.5em minus 0.4em\relax IEEE, 2020, pp. 8229--8233.

\bibitem{jia2022cvss}
Y.~Jia, M.~T. Ramanovich, Q.~Wang, and H.~Zen, ``Cvss corpus and massively multilingual speech-to-speech translation,'' \emph{arXiv preprint arXiv:2201.03713}, 2022.

\bibitem{wang2021covost}
C.~Wang, A.~Wu, J.~Gu, and J.~Pino, ``Covost 2 and massively multilingual speech translation.'' in \emph{Interspeech}, 2021, pp. 2247--2251.

\bibitem{bapna2022mslam}
A.~Bapna, C.~Cherry, Y.~Zhang, Y.~Jia, M.~Johnson, Y.~Cheng, S.~Khanuja, J.~Riesa, and A.~Conneau, ``mslam: Massively multilingual joint pre-training for speech and text,'' \emph{arXiv preprint arXiv:2202.01374}, 2022.

\bibitem{jia2021png}
Y.~Jia, H.~Zen, J.~Shen, Y.~Zhang, and Y.~Wu, ``Png bert: Augmented bert on phonemes and graphemes for neural tts,'' \emph{arXiv preprint arXiv:2103.15060}, 2021.

\bibitem{afouras2018lrs3}
T.~Afouras, J.~S. Chung, and A.~Zisserman, ``Lrs3-ted: a large-scale dataset for visual speech recognition,'' \emph{arXiv preprint arXiv:1809.00496}, 2018.

\end{thebibliography}

\end{document}